\documentclass[10pt,a4paper]{article}
\newcommand{\arxivpreprint}{}
\usepackage[a4paper,margin=2.5cm]{geometry}
\usepackage[T1]{fontenc}
\usepackage[utf8]{inputenc}
\usepackage{lmodern}
\usepackage{xcolor}
\usepackage{microtype}
\usepackage[authoryear,round]{natbib}
\setcitestyle{citesep={;},aysep={,},yysep={;}}

\usepackage{amsmath,amsfonts,bm}

\def\eqref#1{equation~\ref{#1}}

\def\1{\bm{1}}

\DeclareMathAlphabet{\mathsfit}{\encodingdefault}{\sfdefault}{m}{sl}
\SetMathAlphabet{\mathsfit}{bold}{\encodingdefault}{\sfdefault}{bx}{n}

\newcommand{\E}{\mathbb{E}}

\usepackage[hyperfootnotes=false]{hyperref}
\usepackage{url}
\usepackage{graphicx}
\setkeys{Gin}{height=0.78\textheight,keepaspectratio}
\usepackage{booktabs}
\usepackage{subcaption}
\usepackage{listings}
\usepackage{prompt_templates}
\usepackage{float}
\usepackage{placeins}
\usepackage{needspace}
\usepackage{etoolbox}
\usepackage{wrapfig}
\usepackage{enumitem}
\usepackage{array}
\hypersetup{
  colorlinks=true,
  citecolor=[RGB]{50,100,170},
  linkcolor=[RGB]{50,100,170},
  urlcolor=[RGB]{50,100,170},
  breaklinks=true,
  pdfborder={0 0 0}
}

\usepackage[nameinlink,capitalize]{cleveref}
\newcommand{\pref}[1]{\cref{#1}}
\newcommand{\savehyperref}[2]{\texorpdfstring{\hyperref[#1]{#2}}{#2}}

\crefformat{equation}{#2Eq. (#1)#3}
\Crefformat{equation}{#2Eq. (#1)#3}
\Crefformat{figure}{#2Figure #1#3}
\Crefname{subsubsection}{Section}{Sections}
\crefformat{subsubsection}{#2Section #1#3}
\Crefformat{subsubsection}{#2Section #1#3}

\renewcommand{\eqref}[1]{%
  \texorpdfstring{\hyperref[#1]{(\ref*{#1})}}{(\ref*{#1})}}
\newcommand{\sref}[1]{%
  \texorpdfstring{\hyperref[#1]{Section \ref*{#1}}}{Section \ref*{#1}}}
\newcommand{\aref}[1]{%
  \texorpdfstring{\hyperref[#1]{Appendix \ref*{#1}}}{Appendix \ref*{#1}}}

\usepackage{crossreftools}
\setlist{topsep=4pt,itemsep=2pt,parsep=0pt}

\makeatletter
\renewcommand\section{\@startsection{section}{1}{\z@}%
  {-2.0ex plus -0.5ex minus -.2ex}{1.5ex plus .3ex minus .2ex}%
  {\large\bfseries\raggedright}}
\renewcommand\subsection{\@startsection{subsection}{2}{\z@}%
  {-1.8ex plus -0.5ex minus -.2ex}{.8ex plus .2ex}%
  {\normalsize\bfseries\raggedright}}
\renewcommand\subsubsection{\@startsection{subsubsection}{3}{\z@}%
  {-1.5ex plus -0.5ex minus -.2ex}{.5ex plus .2ex}%
  {\normalsize\bfseries\raggedright}}
\makeatother

\renewcommand{\enlargethispage}[1]{}

\let\arxivsubsection\subsection
\RenewDocumentCommand{\subsection}{s o m}{%
  \IfBooleanF{#1}{%
    \ifstrequal{#3}{Qualitative analysis of learned harnesses}{\Needspace{24\baselineskip}}{}}%
  \IfBooleanTF{#1}{\arxivsubsection*{#3}}{%
    \IfNoValueTF{#2}{\arxivsubsection{#3}}{\arxivsubsection[#2]{#3}}}%
}

\newif\ifdraftnotes
\draftnotestrue

\title{Harness Learning Enables \\ Generalizable Test-Time Adaptation}
\author{Alvin Zhang\thanks{Equal contribution.}\thanks{Project lead.}, Xuecheng Liu\textsuperscript{*}, Zixuan Wang\textsuperscript{*}\\[3pt]Fahim Tajwar, Daman Arora, Ruslan Salakhutdinov\\[3pt]Daniel Khashabi, Yuda Song\thanks{Equal advising.}, Andrea Zanette\textsuperscript{\ddag}}
\date{}
\hypersetup{pdftitle={Harness Learning Enables Generalizable Test-Time Adaptation},pdfauthor={Alvin Zhang, Xuecheng Liu, Zixuan Wang, Fahim Tajwar, Daman Arora, Ruslan Salakhutdinov, Daniel Khashabi, Yuda Song, Andrea Zanette}}

\makeatletter
\renewcommand{\maketitle}{%
  \begingroup
  \setlength{\parskip}{0pt}%
  \renewcommand{\thefootnote}{\fnsymbol{footnote}}%
  \noindent\makebox[\textwidth]{%
    \raisebox{-0.5\height}{\includegraphics[width=0.35\textwidth]{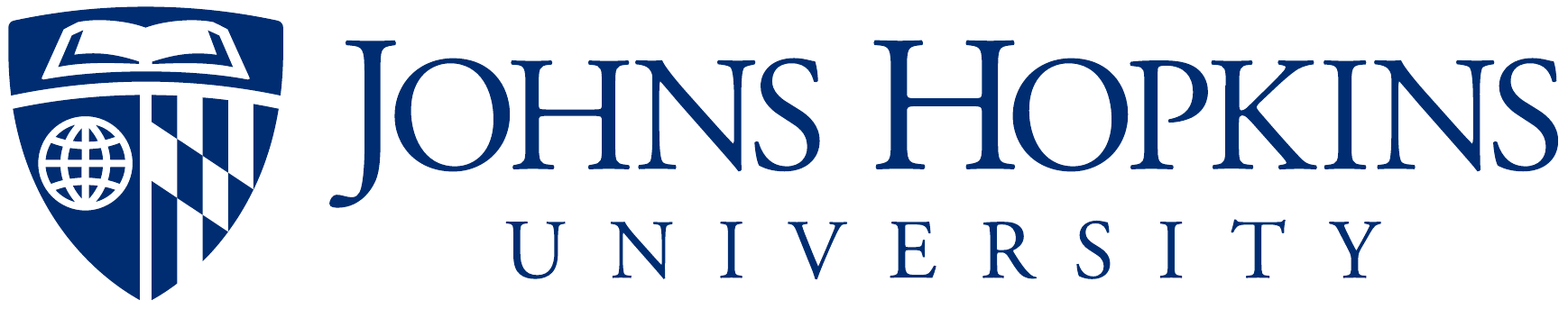}}%
    \hfill
    \raisebox{-0.5\height}{\includegraphics[width=0.51\textwidth]{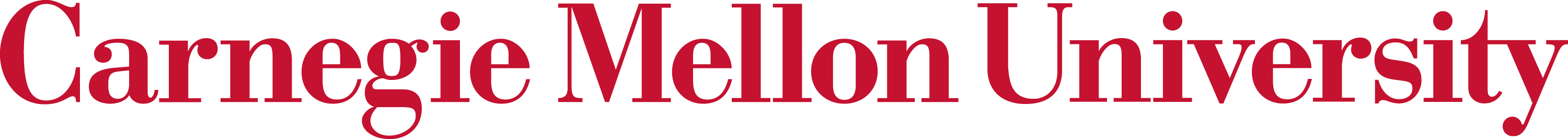}}}%
  \par\vspace{16pt}%
  \noindent{\color{black}\hrule height 1pt}\vspace{.25in}%
  {\centering\LARGE\bfseries\@title\par}%
  \ifx\@author\@empty\else
    \vspace{12pt}%
    {\centering\normalsize\@author\par}%
  \fi
  \@thanks
  \par\vspace{.3in}%
  \endgroup
  \setcounter{footnote}{0}%
  \thispagestyle{plain}%
}
\makeatother
\renewenvironment{abstract}{%
  \par\begin{center}\large\bfseries Abstract\end{center}%
  \begin{list}{}{\setlength{\leftmargin}{.6cm}\setlength{\rightmargin}{.6cm}}%
  \item[]\ignorespaces
}{\unskip\end{list}}

\begin{document}

\maketitle

\begin{abstract}
A language-model agent is jointly defined by its model and its harness, the executable program that organizes model calls, tool use, and information flow. Because different tasks call for different ways of organizing these operations, the harness needs to be adapted using feedback from the task at hand. We introduce \emph{harness learning}, which trains a proposer model to revise a solver's harness using execution feedback. We formulate this process as meta-learning over executable programs, with harness revisions playing the role of weight updates in gradient-based adaptation. We train the proposer with reinforcement learning, using the task performance of revised harnesses as the reward. At test time, the proposer uses feedback from successive executions on a new task to refine the harness, without performing any parameter-space update. Experiments on reasoning and multi-hop question answering show that harness learning improves revision quality and that the ability to adapt at test time transfers to unseen tasks. Policies trained on individual revisions can continue improving harnesses over multiple rounds, while the benefits of training on revision sequences vary across settings. These findings suggest a path towards continually learning agents that turn accumulated experience into generalizable improvements.
\end{abstract}

\section{Introduction}
\label{intro}

A \emph{harness} is the executable program that organizes calls to a foundation model and interactions with tools \citep{yang2024sweagent,khattab2023dspycompilingdeclarativelanguage}. It determines what the model sees, which tools it can use, and how execution proceeds. Because these choices strongly shape performance \citep{shinn2023reflexionlanguageagentsverbal,yang2024sweagent}, adapting the harness provides a way to improve how a model uses its capabilities without changing its parameters \citep{zhang2025aflowautomatingagenticworkflow,lou2026autoharnessimprovingllmagents}.

Execution feedback can guide choices about model calls, tool use, and intermediate outputs, but finding effective harness revisions often requires repeated experimentation at test time, and revisions that work on one task may not transfer to another. We therefore ask \emph{whether a model can learn to revise harnesses from execution feedback in a way that generalizes to new tasks}. We call this problem \emph{harness learning} and view it as \emph{meta-learning} over executable programs \citep{duan2016rl2,finn2017maml}. The outer loop trains a revision model, which we call the \emph{proposer}, from execution outcomes, and the inner loop uses the proposer to adapt a harness from execution feedback. Training therefore learns a reusable revision procedure that can transfer to unseen tasks without updating model parameters at test time \citep{baxter2000model,andrychowicz2016learning}.

\begin{figure}[t]
    \centering
    \includegraphics[width=\textwidth]{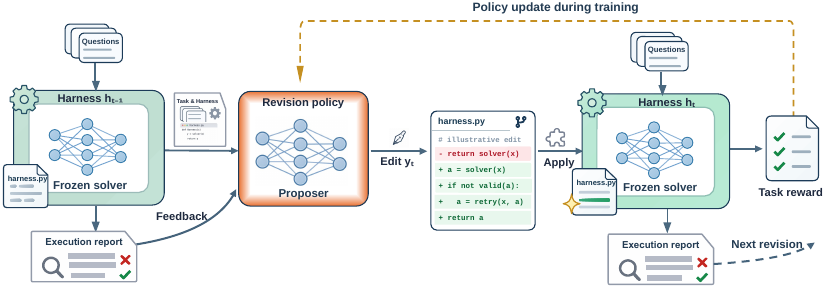}
    \caption{\textbf{Harness learning.} The proposer reads the task description, the current harness $h_{t-1}$, and its execution report, and writes a code edit $y_t$, here adding answer validation and a retry. Running the revised harness $h_t$ yields a reward and feedback for the next round. Training updates only the proposer, and at test time both models are frozen. \pref{fig:pipeline} details training.}
    \label{fig:teaser}
    \vspace{-14pt}
\end{figure}

We implement harness learning by training the proposer to edit executable harness code (\pref{fig:teaser}). Given a task description, the current harness, and an execution report, the proposer generates a code change. We train the proposer with reinforcement learning, using the performance of the resulting harness as the reward, and optionally initialize it with supervised fine-tuning. At test time, the same proposer can be applied repeatedly, using feedback from each execution to guide the next revision.

We evaluate harness learning on reasoning and multi-hop question answering. On Reasoning Gym \citep{stojanovski2025reasoninggym}, training improves revision quality on task families excluded from both supervised and reinforcement learning, and in single-step revision the trained 4B proposer outperforms the 35B teacher on average. In multi-hop question answering, a proposer trained on HotpotQA \citep{yang2018hotpotqa} transfers to MuSiQue \citep{trivedi2022musique} and 2WikiMultihopQA \citep{ho2020twowiki}. In both settings, trained proposers can continue improving harnesses over successive rounds of test-time adaptation. In QA, even a proposer trained only on seed-harness inputs continues improving revised harnesses, suggesting generalization across adaptation contexts and tasks.

We make the following contributions.

\noindent\textbf{Learning to adapt executable harnesses at test time.}
We introduce harness learning as meta-learning over executable programs: a model learns from execution outcomes how to revise the harness that organizes model calls and tool use. At test time, the model uses execution feedback to adapt harnesses for new tasks without any parameter updates. 

\noindent\textbf{Generalization to unseen tasks.}
We show that the capability of test-time adaptation via harness revision transfers to reasoning tasks and question-answering benchmarks that are unseen during training.

\noindent\textbf{Iterative test-time adaptation.}
We show that proposers trained on individual revisions can continue improving harnesses over successive rounds of test-time adaptation. We analyze how training changes the reliability of the revision, the structure of the auxiliary, and the use of generated tools.

\section{Related Work}
\label{related_works}

\enlargethispage{-9pt}

\paragraph{Harness optimization.} Automated agent design optimizes language-model pipelines and workflows \citep{khattab2023dspycompilingdeclarativelanguage,zhang2025aflowautomatingagenticworkflow}, with harness-specific methods using execution feedback to revise the programs that govern agent behavior \citep{lee2026metaharnessendtoendoptimizationmodel,lou2026autoharnessimprovingllmagents}. Concurrent work also trains dedicated models for harness construction and revision \citep{shao2026harnessr1learningeditexecutable,zhang2026jitagentscalingharnessintelligence}. Harness-R1 trains an editor through supervised initialization and reinforcement learning with a frozen solver, a training recipe similar to ours. JIT-Agent learns task-conditioned harness generation, repair, and evolution. Our study frames harness revision as a learned adaptation procedure and examines whether successive rounds of execution feedback improve performance on task families excluded from training.

\paragraph{Meta-learning.} Meta-learning uses experience across tasks to learn an initialization or an adaptation procedure \citep{duan2016rl2,finn2017maml,hospedales2021metalearning}. AdaptFlow learns a shared workflow initialization \citep{zhu2025adaptflowadaptiveworkflowoptimization}, and The Last Harness proposes an outer loop over harness-evolution blueprints \citep{seong2026harnessyoullbuild}. Our formulation treats executable harness code as the adapted object and trains the proposer as the adaptation rule. Treating harness code as the adapted object lets the inner loop change control flow, tool interfaces, and the sequence of model calls, and the proposer learns from execution feedback how to make these discrete changes. At test time, both proposer and solver remain frozen, and graded execution feedback on development questions guides successive harness revisions. TTHE also adapts executable harnesses with frozen models, using unlabeled execution traces as feedback \citep{nie2026tthe}. \aref{app:related_polished} discusses broader adaptation methods and evaluation protocols. 

\section{Preliminaries}
\subsection{Harness}

Given a language model, a \emph{harness} is the executable program that orchestrates the model's calls throughout a task. The harness constructs the context for each model call, controls when calls occur, executes tool requests, and incorporates environmental feedback to support iterative problem solving. Harness design determines how a model interacts with its environment. For example, a coding-agent harness can expose tools for inspecting files, editing code, and running tests, then return the resulting observations to the model \citep{yang2024sweagent}. A harness can be changed while the model's parameters stay fixed.

\subsection{Reinforcement learning with verifiable rewards}
\label{sec:prelim_rl}
A \emph{policy} $\pi_\theta(y\mid x)$ is an autoregressive language model that maps a context $x$ to a token sequence $y=(y_1,\dots,y_{|y|})$. In reinforcement learning with verifiable rewards, a programmatic grader evaluates responses or their execution outcomes. Training maximizes the expected reward over a prompt distribution $\mathcal{D}$,
\begin{equation}
\mathcal{J}(\theta)=\E_{x\sim\mathcal{D}}\,\E_{y\sim\pi_\theta(\cdot\mid x)}\big[r(x,y)\big].
\label{eq:rl_objective}
\end{equation}

\paragraph{GRPO.} Group relative policy optimization \citep{shao2024deepseekmath} normalizes rewards within a group of responses $\{y_i\}_{i=1}^{G}$ sampled from $\pi_{\theta_{\mathrm{old}}}$ for the same context. With rewards $r_i=r(x,y_i)$, each token of $y_i$ receives the group-normalized advantage $\hat{A}_i=(r_i-\operatorname{mean}_j r_j)/(\operatorname{std}_j r_j+\delta)$. Let $\omega_{i,k}(\theta)=\pi_\theta(y_{i,k}\mid x,y_{i,<k})/\pi_{\theta_{\mathrm{old}}}(y_{i,k}\mid x,y_{i,<k})$ denote the token-level importance ratio. The clipped surrogate objective is
\begin{equation}
\resizebox{0.90\linewidth}{!}{$\displaystyle \mathcal{J}(\theta)=\E_{x\sim\mathcal{D}}\,\E_{\{y_i\}\sim\pi_{\theta_{\mathrm{old}}}(\cdot\mid x)}\left[\frac{1}{\sum_{j=1}^{G}|y_j|}\sum_{i=1}^{G}\sum_{k=1}^{|y_i|}\min\!\Big(\omega_{i,k}(\theta)\hat{A}_i,\ \operatorname{clip}(\omega_{i,k}(\theta),1-\epsilon,1+\epsilon)\,\hat{A}_i\Big)\right]$}
\label{eq:grpo}
\end{equation}

\section{Harness Learning as Meta-Learning}
\label{method}

Harness learning trains a proposer to adapt the executable harness of a fixed solver. The proposer learns from revision outcomes during training and uses execution feedback to revise harnesses on new tasks at test time. We first define how a harness is evaluated, then describe the adaptation loop and the procedure used to train its revision rule.

\subsection{Tasks, harnesses, and execution feedback}
\label{sec:setup}

A task provides a description, a set of questions, and a grader that assigns scores in $[0,1]$. The task is the unit of harness adaptation. In our experiments, a task is either a Reasoning Gym family, such as \texttt{maze} or \texttt{sudoku}, whose questions share a generator and a grader, or a multi-hop QA benchmark such as HotpotQA. A harness $h$ answers questions by organizing calls to the solver, tool use, and direct computation. We keep the solver fixed and omit it from the notation.

\paragraph{Harness quality.}
Let $\operatorname{score}(h,q)$ denote the observed grader score from executing harness $h$ on question $q$. We evaluate a harness on a finite question set $Q$ using its mean score,
\begin{equation}
J(h;Q)=\frac{1}{|Q|}\sum_{q\in Q}\operatorname{score}(h,q).
\label{eq:harness_quality}
\end{equation}
The score is empirical and can vary across executions when the solver is stochastic. Execution also produces traces, which we summarize together with task outcomes in an \emph{execution report}. \aref{app:prompts} gives the report formats.

\paragraph{Question sets.}
We divide the full question set into subsets that serve different purposes. We use $Q^{\mathrm{fb}}$ to construct execution reports, $Q^{\mathrm{score}}$ to score candidate revisions, and $Q^{\mathrm{eval}}$ to measure held-out performance. Evaluation questions are disjoint from all questions used for feedback or candidate selection during adaptation. Feedback and scoring sets may coincide when the evaluation protocol uses the same development questions for both roles. Round subscripts indicate the sets used at a particular revision round. \looseness=-1

\subsection{Harness adaptation and the meta-learning objective}
\label{sec:inference}

The proposer $\pi_\theta$ defines the learned revision rule. At test time, its parameters $\theta$ and the solver's parameters remain fixed while the harness changes. Starting from a seed harness $h_0$, adaptation proceeds for $T$ rounds.

At round $t$, the proposer input $x_t$ contains the task description, the current harness $h_{t-1}$, and a report from executing that harness on $Q_t^{\mathrm{fb}}$. The input may also include selected information from earlier rounds. The proposer samples $G$ responses, applies their edits, and retains a harness according to the selection protocol,
\begin{equation}
\begin{aligned}
y_i &\sim \pi_\theta(\cdot\mid x_t), && i=1,\ldots,G,\\
h_i' &= \operatorname{Apply}(h_{t-1},y_i),\\
h_t &= \operatorname{Select}\!\left(h_{t-1},\{h_i'\}_{i=1}^{G};Q_t^{\mathrm{score}}\right).
\end{aligned}
\label{eq:revision_round}
\end{equation}
The index $i$ identifies candidates within the current round. $\operatorname{Apply}$ extracts code edits from the generated response and applies them to the parent harness, marking proposals invalid when their edits cannot be parsed or applied. In our structured interface, an edit identifies a span of existing code and supplies its replacement.

$\operatorname{Select}$ evaluates candidates on the scoring questions and applies the protocol's validity and retention rules. For example, our Reasoning Gym protocol samples one candidate per round ($G=1$), uses the same development questions for feedback and scoring, and retains the candidate only if its development score exceeds that of the current harness (\aref{app:revision_selection}). Our QA protocol samples eight candidates, assigns a score of zero to those that fail to parse or run, and advances to the highest-scoring candidate even when it scores below its parent (\aref{app:qa_eval_details}). Executing the retained harness on the next round's feedback questions provides the report for $x_{t+1}$.

\paragraph{Learning to adapt.}
For a fixed proposer, the revision loop maps a seed harness $h_0$ to an adapted harness $h_T$ and forms the inner loop of meta-learning. The outer-loop goal is to learn proposer parameters that improve the held-out performance of the adapted harness,
\begin{equation}
\max_\theta\ \mathbb{E}\!\left[J(h_T;Q^{\mathrm{eval}})\right].
\label{eq:outer}
\end{equation}
Here $h_T$ is produced by \pref{eq:revision_round} using $\pi_\theta$. The expectation averages over the training-task distribution, sampled questions, proposed revisions, and stochastic harness executions. Training can draw from one or several tasks, as specified in \sref{sec:settings}.

Our training procedure uses the immediate performance of individual revisions as a surrogate for this final-harness objective. Each update assigns credit from the candidate's own outcome. Training on successive revision states changes the inputs encountered by the proposer while retaining this local objective.

\subsection{Training the proposer}
\label{sec:rl}

Training updates only the proposer. Each revision input $x$ combines a training task, a parent harness, and an execution report, so the input distribution $\mathcal D$ in \sref{sec:prelim_rl} is a distribution over these revision contexts.

\paragraph{Optional supervised initialization.}
\label{sec:sft}
We optionally initialize the proposer with successful teacher revisions. For each context used to collect SFT data, a teacher receives the seed harness and its execution report and proposes edits. We execute the resulting harnesses and retain revisions that improve on the seed and pass the quality filters in \aref{app:sft_details}. Supervised fine-tuning maximizes the likelihood of the retained teacher responses, providing an initialization for learning from the outcomes of the proposer's own revisions.

\begin{figure}[t]
    \centering
    \includegraphics[width=\textwidth]{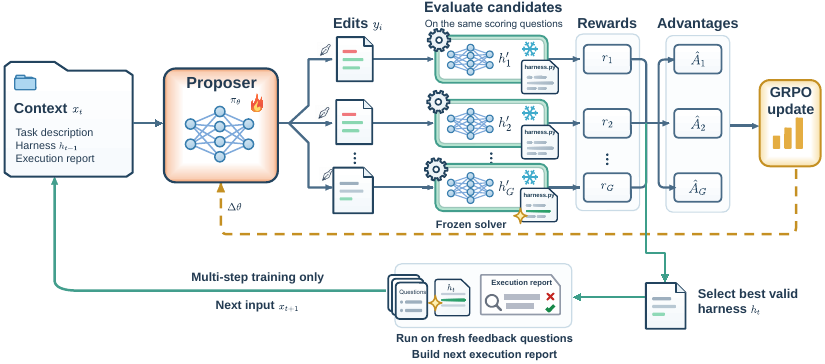}
    \caption{\textbf{Reinforcement learning for harness revision.} Given input $x_t$, the proposer samples $G$ edits to the parent harness $h_{t-1}$. Each candidate $h_i'$ runs on the same scoring questions, which gives its reward $r_i$ and a group-normalized advantage. For successive revision states, a selected harness $h_t$ runs on fresh feedback questions, disjoint from the scoring questions, to form the next input $x_{t+1}$.}
    \label{fig:pipeline}
\end{figure}

\paragraph{Reinforcement learning from revision outcomes.}
For each input, we sample candidate revisions as in \pref{eq:revision_round} and evaluate them on a common scoring set from the same task. During training, scoring questions are disjoint from those used to construct the input's execution report. An evaluable candidate receives reward
\begin{equation}
r_i=J(h_i';Q^{\mathrm{score}})+v_i,
\label{eq:reward}
\end{equation}
where $v_i$ is an auxiliary reward for edit validity and successful execution. These rewards determine the group-normalized advantages used by GRPO (\sref{sec:prelim_rl}). \aref{app:rl_details} specifies the validity terms for Reasoning Gym, where a failed edit receives only the validity credit accumulated before the failure and explicit no-change responses are scored by executing the parent. \pref{fig:pipeline} illustrates the update.

\paragraph{Training on successive revision states.}
Single-step training treats each parent harness and report as an independent revision context. Training on successive revisions constructs additional contexts from selected candidate harnesses and their refreshed execution reports. For Reasoning Gym, we refresh these states between offline training phases; for QA, we generate revision sequences online. Each state uses the same immediate-reward update, so both procedures train the proposer to improve the harness presented in its current input. \sref{sec:settings} and \aref{app:multistep_details} describe the configurations. We call the resulting configurations Single-step RL and Multistep RL. Either trained proposer can be applied over multiple revision rounds at test time.

\section{Experiments}

\begin{figure}[t]
    \centering
    \includegraphics[width=\textwidth]{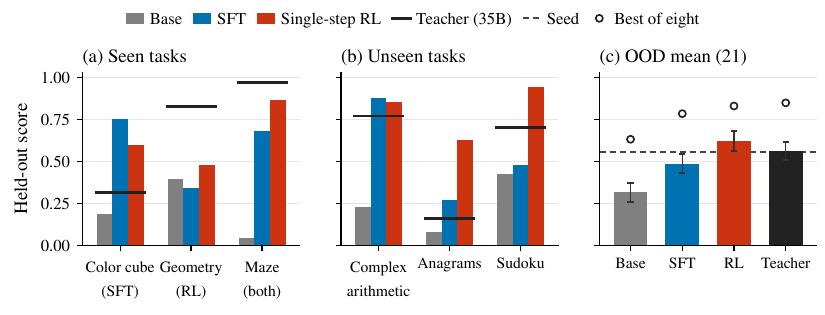}
    \captionsetup{skip=4pt}
    \caption{\textbf{Single-step harness revision.} Held-out scores on format-varied questions for (a) three seen and (b) three unseen families. Bars average eight proposals, and black lines mark the teacher. (c) Mean over 21 unseen families with one standard error. Circles mark oracle best-of-eight scores, selected on held-out scores, and the dashed line marks the seed. Full results are in \pref{tab:single_step_families}.}
    \label{fig:single_step}
    \vspace{-14pt}
\end{figure}

\begin{figure}[t]
    \centering
    \includegraphics[width=\textwidth]{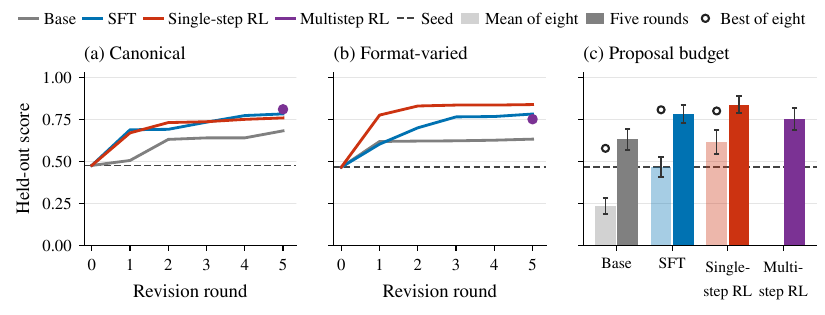}
    \caption{\textbf{Multistep harness revision on unseen families.} (a) Canonical and (b) format-varied held-out scores over five rounds, averaged over 15 unseen families. Harnesses are retained by development score on separate questions that also supply feedback, and Multistep RL is shown at round five. (c) Eight independent proposals and five sequential revisions on format-varied questions. Hollow circles mark oracle best-of-eight scores, dashed lines mark the seed, and error bars show one standard error across families.}
    \label{fig:multistep_ood}
    \vspace{-14pt}
\end{figure}

To evaluate harness learning, we organize our experiments around three questions:
\begin{itemize}[leftmargin=*]
    \item Do learned revisions generalize to tasks unseen during training?
    \item Does successive test-time revision improve harnesses beyond a single revision?
    \item Does training on revision sequences improve adaptation over training on individual revisions?
\end{itemize}
We study these questions in two settings with separately trained proposers. Reasoning Gym tests generalization across diverse synthetic task families from a minimal seed harness that makes one solver call. Multi-hop question answering tests whether learned revisions improve an existing retrieval workflow.

\subsection{Experimental setup}
\label{sec:settings}
\pref{tab:settings} summarizes the two settings, and \savehyperref{app:rg_training}{Appendices~\ref*{app:rg_training}} and~\ref{app:qa_details} give the full protocols. \emph{Unseen} (OOD) tasks are excluded from the proposer's supervised fine-tuning and reinforcement learning. Seed denotes the unrevised seed harness, and Base denotes revisions generated by the base proposer before training. SFT, Single-step RL, and Multistep RL denote training configurations, and each resulting proposer can perform single-step or multistep revision at test time. For Reasoning Gym, \emph{canonical} questions use the benchmark's standard format, while \emph{format-varied} questions present the same tasks in alternative formats.

\begin{table}[!ht]
\centering
\small
\caption{\textbf{Experimental settings.} Oracle best-of-$N$ statistics select candidates by held-out score.}
\label{tab:settings}
\begin{tabular}{@{}>{\raggedright\arraybackslash}p{0.10\linewidth}>{\raggedright\arraybackslash}p{0.42\linewidth}>{\raggedright\arraybackslash}p{0.42\linewidth}@{}}
\toprule
 & Reasoning Gym & Multi-hop QA \\
\midrule
Models & Qwen3.5-4B proposer and solver & Qwen3-4B proposer, Qwen3-8B solver \\
Training & SFT on 35B-teacher revisions, then RL; Multistep RL continues from Single-step RL & RL from Base; Multistep RL trains on online revision runs \\
Tasks & 21 SFT and 5 RL families; 21 unseen families & Train on HotpotQA; MuSiQue and 2WikiMultihopQA unseen \\
Revision & 8 independent proposals on format-varied questions; 5 rounds of one proposal, kept only if development score improves & 80 independent proposals; 4 runs of 10 rounds, 8 proposals per round \\
Reported & Held-out score; best of eight as an oracle & Held-out exact match; best of $N$ as an oracle \\
\bottomrule
\end{tabular}
\end{table}

\subsection{Learning transferable revisions on Reasoning Gym}
\label{sec:results_rg}
\enlargethispage{-9pt}

\begin{figure}[t]
    \centering
    \includegraphics[width=\textwidth]{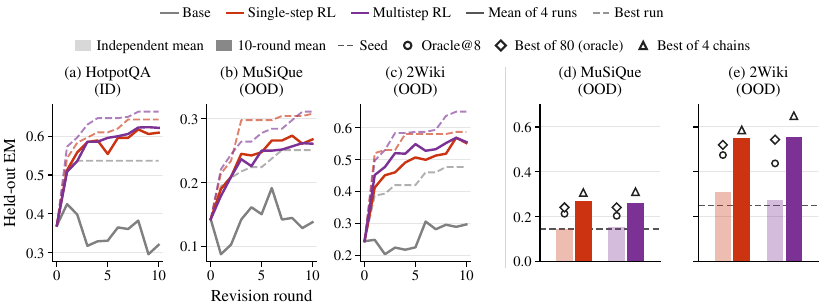}
    \caption{\textbf{Independent and sequential QA revision.} (a,b,c) Held-out exact match over ten rounds. Solid lines average four runs, and dashed lines show the running maximum across runs. (d,e) On the unseen benchmarks, light bars average 80 independent revisions and dark bars average the final scores of four ten-round runs with the same 80-proposal budget. Circles, diamonds, and triangles mark the oracle best of 8, the best of 80, and the best over all rounds and runs. Dashed lines mark the seed. \aref{app:qa_single_step} gives the protocol.}
    \label{fig:multihop_multistep}
    \label{fig:hotpotqa}
    \label{fig:musique}
    \vspace{-14pt}
\end{figure}

\paragraph{Generalization of individual revisions.}
Both training stages improve mean single-revision scores on 21 unseen families, from 0.32 for Base to 0.62 after RL (\pref{fig:single_step}, \aref{app:protocol}). The improvement appears in the mean over all proposals, so it does not depend on selecting the best candidate. After RL, the average proposal already exceeds the seed harness, while Base and SFT proposals fall below it on average.

\paragraph{Comparison with the teacher.}
On the 21 unseen families, the trained 4B proposer's revisions score above those of its 35B teacher on average (0.62 versus 0.56) under the same revision protocol. The teacher achieves higher scores under oracle best-of-eight selection. Learning from execution outcomes after supervised initialization can therefore raise average proposal quality above the teacher's. \aref{app:protocol} details the comparison, including the treatment of failed edits and zero-scoring harnesses.

\paragraph{Gains across revision rounds.}
On format-varied questions, Single-step RL obtains most of its five-round gain in the first revision, while SFT improves more gradually across rounds (\pref{fig:multistep_ood}b). On canonical questions, the two proposers improve on similar schedules (\pref{fig:multistep_ood}a). \sref{sec:analysis_meta} discusses why later revisions add less.

\paragraph{Sequential versus independent revision.}
With five sequential proposals, the retained harness approaches or exceeds the oracle best of eight independent proposals and exceeds their average (\pref{fig:multistep_ood}c). Building on the current best harness therefore uses a small proposal budget effectively. \aref{app:multistep} gives in-domain and per-family results.

\subsection{Adapting retrieval workflows on unseen QA benchmarks}
\label{sec:results_qa}
\enlargethispage{-9pt}

\begin{figure}[t]
    \centering
    \includegraphics[width=\textwidth,trim=0 6bp 0 0,clip]{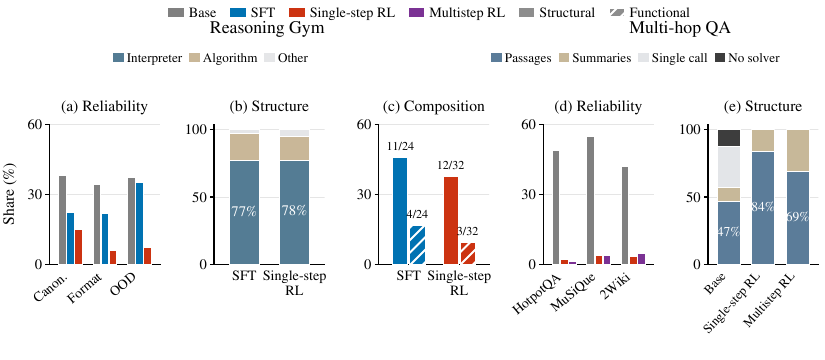}
    \captionsetup{skip=4pt}
    \caption{\textbf{Proposal reliability and harness structure.} (a) Failed or zero-scoring proposals in Reasoning Gym (15 canonical, 12 format-varied, and 5 unseen families). (b) Harness classes on five unseen families without a composition directive. (c) Composition under a helper-tool directive on four unseen families. Structural composition places helpers in the solver's tool loop, and functional composition also requires a helper to contribute to the answer. (d) Failed or zero-scoring proposals in QA (320 candidates per benchmark and method). (e) Structures of retained QA harnesses over ten rounds, pooled over three benchmarks ($n=120$ per method), where Passages and Summaries denote multi-hop retrieval from passages or solver summaries. Counts are in \pref{tab:validity,tab:adjudication}.}
    \label{fig:validity}
    \label{fig:reliability_structure}
    \label{fig:qa_analysis}
    \vspace{-14pt}
\end{figure}

\paragraph{Progress over successive revisions.}
Both RL proposers continue to improve beyond the first revision and finish above Base on all three benchmarks, including the two unseen ones, while Base ends below the seed on HotpotQA and MuSiQue (\pref{fig:multihop_multistep}a--c). Improvement also persists over more rounds than on Reasoning Gym, and the first revision accounts for only one third to two thirds of the ten-round gain. \sref{sec:analysis_meta} discusses this difference.

\paragraph{Sequential versus independent revision.}
On MuSiQue, Single-step RL averages 0.15 exact match for independent revisions, only marginally above the seed (\pref{tab:qa_single_step}), versus 0.27 after ten revision rounds. On both unseen benchmarks, the average final run of each RL proposer exceeds even the oracle best of 80 independent proposals (\pref{fig:multihop_multistep}d--e). Average improvement from the seed therefore does not fully capture a proposer's ability to improve harnesses through repeated adaptation. Each run uses 80 proposals, so the comparison matches the number of proposals per run and leaves total compute and feedback unmatched.

\subsection{Effect of training on revision sequences}
\label{sec:results_seq}
Training on successive revision contexts with immediate rewards does not consistently improve on training on individual revisions. On Reasoning Gym, Multistep RL ends above Single-step RL on canonical questions and below it on format-varied questions (\pref{fig:multistep_ood}a,b), and training on fixed later-round states also shows no gain (\aref{app:later_round_states}). In QA, the two configurations reach nearly identical final scores (\pref{fig:multihop_multistep}a--c). In both settings, proposers trained only on individual revisions already improve harnesses over successive rounds, which suggests that iterative improvement can emerge from single-revision training.

\section{Analysis and Discussion}
\label{sec:analysis}
We examine how training changes the reliability and structure of proposals, and what these changes suggest about repeated revision.

\subsection{Training improves proposal reliability}
\label{sec:analysis_sft}

\enlargethispage{-9pt}
Training reduces the fraction of proposals that fail to produce a runnable harness or score zero on the development questions (\pref{fig:reliability_structure}a,d; \pref{tab:validity}). On unseen Reasoning Gym families, most of the reduction comes from RL, and SFT alone leaves the rate nearly unchanged. In QA, RL from the base proposer produces a similar reduction. A trained proposer therefore supplies more viable candidates to selection in each round, which may help sustain progress across successive revisions.

\subsection{Qualitative analysis of learned harnesses}
\label{sec:analysis_rl}

\begin{wrapfigure}{R}{0.25\textwidth}
    \centering
    \includegraphics[width=\linewidth]{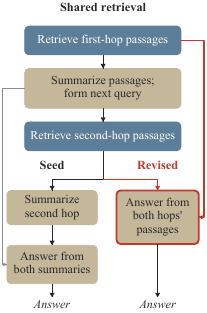}
    \captionsetup{font=small,skip=4pt,justification=raggedright,singlelinecheck=false}
    \caption{\textbf{A learned QA revision.} The answering call reads retrieved passages directly (red), and first-hop summaries still guide retrieval.}
    \label{fig:qa_evidence_revision}
\end{wrapfigure}

On Reasoning Gym, SFT and Single-step RL mostly generate interpreter loops, which let the solver formulate a computation as code and delegate its execution to the runtime (\pref{fig:reliability_structure}b; \pref{tab:adjudication}). Interpreter loops are also the most common structure in the SFT data (48\% of teacher revisions; \aref{app:sft_details}) and remain dominant after RL. Under a composition directive, helpers contribute to the answer in only a minority of the harnesses that integrate them into the solver's tool loop (\pref{fig:reliability_structure}c), and composition rewards do not yield consistent gains (\aref{app:composition}). Unsuccessful composed harnesses often pair a correct helper algorithm with failures in question parsing, tool communication, or answer extraction (\pref{tab:composition_failures}), so a generated component contributes only when these interfaces also work.

In QA, RL preserves multi-hop retrieval, while Base sometimes reduces the harness to a single solver call or removes the solver (\pref{fig:reliability_structure}e). A common RL revision keeps summaries for guiding the next retrieval and passes retrieved passages directly to the answering call (\pref{fig:qa_evidence_revision}), which separates the evidence needed to formulate a query from the evidence needed to answer. Other revisions add a third retrieval hop or search again when a draft answer is absent from the retrieved passages (\pref{fig:qa_flowcharts}). In both settings, the trained proposers change how the harness organizes computation and information. Their transfer to unseen tasks suggests that they learn a reusable adaptation capability from execution outcomes.

\subsection{Repeated revision}
\label{sec:analysis_meta}

\enlargethispage{-9pt}
The seed harness may explain why QA gains accumulate over more rounds than Reasoning Gym gains. On Reasoning Gym, single-step revisions commonly introduce interpreter loops. In the composition experiments, a correct helper algorithm may still fail to contribute through the surrounding parsing and tool interfaces, which is one obstacle to further gains. The QA seed already contains a retrieval workflow whose depth, evidence access, and search triggers can each be refined in later rounds. Differences in training and selection procedures prevent attributing this contrast solely to the starting harness.

Repeated revision also tests whether a policy remains useful as its own edits change the adaptation context. In QA, Single-step RL trains on seed-harness inputs yet continues to improve revised harnesses at test time (\pref{fig:multihop_multistep}). This continued improvement suggests that the learned revision skills of Single-step RL extend to intermediate harnesses and execution reports beyond those encountered in training, as well as to unseen benchmarks.

Our experiments on training with revision sequences change the parent harnesses and execution reports used for training and keep each reward local to a single revision (\sref{sec:rl}). They leave open whether training that assigns credit across rounds would make longer revision chains more effective. Supervised demonstrations of successive revisions, or different distributions of intermediate harnesses and feedback, could yield larger gains from multistep training.

\section{Conclusion}
Harness learning trains a proposer to adapt a frozen solver's executable harness using execution feedback, with both models' parameters fixed at test time. On 21 unseen Reasoning Gym families, SFT followed by RL raises mean single-step revision scores from 0.32 with the base proposer to 0.62, with the trained 4B proposer outperforming its 35B teacher on average. In multi-hop QA, a proposer trained with RL from the base proposer on HotpotQA transfers to unseen benchmarks. Policies trained on individual revisions also improve harnesses over multiple rounds, suggesting that learned revision skills remain useful as the harness changes.

\noindent\textbf{Limitations and future work.} Training uses prescribed revision prompts and a fixed solver, with one teacher for Reasoning Gym supervision. Future work could explore stronger teachers or direct RL from models with prior knowledge of harness design, train tool-using proposers to inspect and test designs, and jointly optimize proposer and solver following concurrent work \citep{ornith2026selfscaffolding,ornith2026selfimprovement}. We emphasize transfer to unseen tasks and longer tool-use trajectories (\aref{app:limitations_future}).

\section*{Acknowledgment}
This work was partially carried out at the Advanced Research Computing at Hopkins (ARCH) core facility (Skipjack), which is supported by the National Science Foundation (NSF) grant number OAC1920103. The authors thank the CMU FLAME center and the CMU Babel Compute Cluster for compute support for this project. This research also used resources of the Oak Ridge Leadership Computing Facility (OLCF) and Argonne Leadership Computing Facility (ALCF)] which are a DOE Office of Science User Facility. This work was supported by an award from the ASCR Leadership Computing Challenge (ALCC) under project ERCAP0034861.
Part of this work was supported by the National Science Foundation under Grant CCF-2106778. YS acknowledge and thank the support of NSF AI Institute for Societal Decision Making AI-SDM grant IIS2229881. FT gratefully acknowledges the support of Qualcomm Innovation Fellowship and Bosch Research and Technology Center.

\clearpage
\bibliography{iclr2027_conference}
\bibliographystyle{plainnat}

\clearpage
\appendix
\setcounter{topnumber}{2}
\section{Extended Related Work}
\label{app:related_polished}

\paragraph{Harness optimization and learned editors.} Harness optimization improves executable agent programs through search or learned editing policies. DSPy optimizes modular language-model pipelines, while AFlow searches over code-based workflows \citep{khattab2023dspycompilingdeclarativelanguage,zhang2025aflowautomatingagenticworkflow}. Meta-Harness and AutoHarness revise harnesses from execution feedback \citep{lee2026metaharnessendtoendoptimizationmodel,lou2026autoharnessimprovingllmagents}, and Ecdysis aggregates failures across instances to guide repairs \citep{yue2026ecdysis}. EvoHarness-RL learns to manage external belief, progress, and experience state during execution \citep{ning2026evoharnessrl}. WHALE \citep{kim2026whalesimplerecipejoint} alternates model weight updates with optimization of the harness with a stronger harness proposer, improving agent performance beyond optimizing a single component alone.

\paragraph{Comparison with contemporary work.} Harness-R1, JIT-Agent, and Ornith are contemporary efforts to learn harness construction or revision. They share our interest in learning from execution outcomes and differ from our work in training signal, adaptation process, and scope of transfer evaluation (\pref{tab:contemporary_harness}).

\paragraph{Harness-R1.} \citet{shao2026harnessr1learningeditexecutable} train a dedicated editor with GPT-5.5 demonstrations followed by GRPO, keeping the target agent frozen. Their SFT, RL, validation, and test partitions contain disjoint instances from the same three benchmarks. They also evaluate patches on instances withheld from adaptation feedback and transfer the editor to unseen target models. Our transfer evaluation excludes entire reasoning families from both training stages and tests a HotpotQA-trained policy on MuSiQue and 2WikiMultihopQA. During Harness-R1 training, rewards come from rerunning the batch that supplied the failure feedback; our scoring questions are disjoint from the feedback questions. Their procedure generates patches without successive refinement of the same patch, whereas we study repeated revision and training on successive revision states. Supervised initialization followed by RL is shared with our Reasoning Gym setting. Our QA proposer starts RL directly from the base proposer, with no teacher-generated revision demonstrations (\sref{sec:settings}).

\paragraph{JIT-Agent.} \citet{zhang2026jitagentscalingharnessintelligence} learn task-conditioned harness generation, repair, and evolution within a four-module protocol. Training combines teacher SFT, execution-based DPO, supervised repair, and Evo-GDPO, which rewards improvements over archive incumbents while accounting for task performance, latency, and cost. Its streaming setting also accumulates execution experience to improve subsequent harnesses. JIT-Agent and our work both study learned adaptation from feedback. JIT-Agent organizes adaptation around task instances and an expanding harness archive; our revision sequences update a current harness evaluated across multiple instances within a task family, with transfer measured on families or benchmarks excluded from training.

\paragraph{Ornith.} Ornith-1.0 jointly trains a policy to generate scaffolds and solve tasks using rewards from solution rollouts \citep{ornith2026selfscaffolding}. Ornith-1.5 extends this loop to jointly optimize task generation, harness generation, and solution rollouts \citep{ornith2026selfimprovement}. These contemporary releases couple improvements in scaffolding with changes to the task-solving policy. Our training updates only the proposer and keeps the solver fixed, so harness improvements are evaluated with unchanged solver parameters. At test time, both models remain fixed while execution feedback guides code revision.

\begin{table}[htbp]
\centering
\small
\setlength{\tabcolsep}{4pt}
\renewcommand{\arraystretch}{1.15}
\caption{\textbf{Contemporary approaches to learned harness adaptation.} Columns give each method's training signal and adaptation process. The accompanying text compares their transfer evaluations.}
\label{tab:contemporary_harness}
\begin{tabular}{@{}p{0.15\linewidth}p{0.36\linewidth}p{0.44\linewidth}@{}}
\toprule
Work & Training & Adaptation \\
\midrule
Harness-R1 & Teacher SFT followed by GRPO & Edit a frozen target agent's executable harness from failure feedback. \\
JIT-Agent & Teacher SFT, execution-based DPO, supervised repair, and Evo-GDPO & Generate, repair, and evolve task-conditioned harnesses using an expanding archive. \\
Ornith & Joint RL for scaffolding and solving; Ornith-1.5 also trains task generation & Improve harness generation together with the task-solving policy. \\
\shortstack[l]{Harness\\Learning} & SFT then RL on Reasoning Gym; direct RL on QA & Revise a current harness with a fixed solver; study transfer and repeated adaptation. \\
\bottomrule
\end{tabular}
\end{table}

\paragraph{Meta-learning.} Meta-learning uses experience across tasks to improve adaptation to new tasks \citep{hospedales2021metalearning}. MAML learns an initialization for gradient-based adaptation \citep{finn2017maml}, while RL$^2$ encodes a reinforcement learning algorithm in a recurrent policy \citep{duan2016rl2}. AdaptFlow learns a shared workflow initialization through language-guided updates \citep{zhu2025adaptflowadaptiveworkflowoptimization}. The Last Harness proposes an outer loop over harness-evolution blueprints \citep{seong2026harnessyoullbuild}, and EvoX jointly evolves solutions and their search strategies \citep{liu2026evoxmetaevolutionautomateddiscovery}. Meta-RL methods also learn to use interaction histories and reflections \citep{yang2026magemetareinforcementlearninglanguage}, convert rubric-based judgments into reusable guidance \citep{li2026rubricemmetarlrubricguidedpolicy}, or train LLMs on diverse interaction trajectories to learn transferable exploration strategies, enabling adaptation to unseen tasks through environmental feedback in context without further parameter updates \citep{tajwar2025traininggenerallycuriousagent}. Our proposer maps a harness and execution feedback to a code revision. Training uses immediate revision rewards as a greedy surrogate for final harness quality (\sref{sec:rl}).

\paragraph{Self-improvement.} Self-improvement methods use feedback to update outputs, memory, code, or model parameters. Self-Refine revises model outputs \citep{madaan2023selfrefineiterativerefinementselffeedback}, and Reflexion stores verbal reflections for later attempts \citep{shinn2023reflexionlanguageagentsverbal}. Feedback can also fail to sustain improvement \citep{jiang2025feedbackfrictionllmsstruggle}. STOP improves the program that proposes code changes \citep{zelikman2024selftaughtoptimizerstoprecursively}, and the Darwin G\"odel Machine evolves agent code through a candidate archive \citep{zhang2026darwingodelmachineopenended}. TTHE adapts harnesses from unlabeled execution traces with fixed model parameters \citep{nie2026tthe}.

Other self-improvement methods update model parameters from self-generated training signals. STaR trains on generated rationales yielding correct answers \citep{zelikman2022starbootstrappingreasoningreasoning}, and Self-Rewarding Language Models construct training preferences from model-generated rewards \citep{yuan2025selfrewardinglanguagemodels}. SEAL learns to generate fine-tuning data and update directives \citep{zweiger2025selfadaptinglanguagemodels}, while SIA combines harness and model-weight updates \citep{hebbar2026siaselfimprovingai}. Harness learning trains a separate proposer from revision outcomes with a fixed solver. At test time, both models remain frozen, and graded development feedback guides harness edits.

\paragraph{Evaluation of harness evolution.} Harness evaluation accounts for solver capability, feedback, and search budgets. \citet{wang2026rethinkingevaluationharnessevolution} examine benchmark overfitting under matched feedback and inference budgets, while \citet{lin2026harnessupdatingharnessbenefit} distinguish generating useful updates from benefiting from them. HarnessCompass studies transfer through constrained edits and richer feedback \citep{zhang2026harnesscompassguidingautomaticharness}, and Evo-Bench evaluates evolution across held-out task suites \citep{huang2026evobenchlanguagemodelsimprove}. Continual-learning evaluations test whether gains persist as tasks arrive \citep{wang2026agentoptimizerscompoundcontinuallearning}. Our experiments fix the solver within each benchmark, separate development feedback from held-out scoring, and report average proposal scores, best-of-eight scores, and successive-revision performance. Evaluations across benchmarks and on families excluded from both training phases test transfer of the learned adaptation procedure.

\paragraph{Continual learning agents.} Continual learning agents must acquire knowledge and skills from ongoing interaction while retaining useful experience. AgentOdyssey evaluates these abilities through procedurally generated, long-horizon text games, with diagnostics for exploration, world knowledge, and episodic memory \citep{zhang2026agentodysseyopenendedlonghorizontext}. Methods address this challenge at both the parameter and program levels. For parametric memory, \citet{zhang2026continuallearningmechanismscompose} show that composing generative replay, self-distillation, and weight regularization with merged LoRA improves retention across sequential fine-tuning tasks. At the program level, Voyager accumulates reusable executable skills \citep{wang2023voyager}, while WorldCoder and Schema construct and revise executable world models from interaction feedback to support planning \citep{tang2024worldcoder,schema2026}. Harness learning complements these directions by learning how to revise the program that organizes model calls and tool use, with model parameters fixed at test time. 

\ifdefined\arxivpreprint\else\newpage\fi
\section{Reasoning Gym Training and Evaluation}
\label{app:rg_training}

\subsection{Evaluation protocol}
\label{app:protocol}
We evaluate revision on training and unseen families, with canonical and format-varied questions (\pref{tab:rg_eval_settings}). Each family has disjoint development and held-out sets. We use development instances for feedback and harness selection, and held-out instances, which remain hidden from the proposer, measure performance.

\begin{table}[htbp]
\centering
\small
\caption{\textbf{Reasoning Gym evaluation settings.} Instance counts, proposal counts, and revision rounds are per task family.}
\label{tab:rg_eval_settings}
\begin{tabular}{@{}lr@{}}
\toprule
Setting & Value \\
\midrule
Development instances & 25 \\
Held-out instances & 75 \\
Independent proposals, single-step & 8 \\
Revision rounds, multistep & 5 \\
Proposals per revision round & 1 \\
Maximum generation attempts per proposal & 3 \\
Response budget per attempt (tokens) & 32,768 \\
\bottomrule
\end{tabular}
\end{table}

\label{app:revision_selection}
Single-step evaluation samples independent revisions of the seed harness from a fixed report. Multistep evaluation revises the best harness retained so far using its development report. At round $t$, the input $x_t$ contains the task description, the retained harness $h^\star_{t-1}$, and its report on $Q^{\mathrm{dev}}$. Applying an edit $y_t\sim\pi_\theta(\cdot\mid x_t)$ produces a candidate $h'_t$, which we retain only if its development score improves,
\begin{equation}
h^\star_t =
\begin{cases}
h'_t, & \text{if } J(h'_t;Q^{\mathrm{dev}})>J(h^\star_{t-1};Q^{\mathrm{dev}}), \\
h^\star_{t-1}, & \text{otherwise.}
\end{cases}
\label{eq:promote}
\end{equation}
Methods share the seed harness and initial report within each comparison. Each subsequent round uses fresh feedback from the retained harness, whose held-out score we measure after every round. \savehyperref{app:prompt_structured}{Appendices~\ref*{app:prompt_structured}} and~\ref{app:prompt_feedback} give the revision template, execution report, and branch instruction.

We score every proposal, including zero-scoring harnesses, and exclude the seed from proposal counts. Edits are parsed only from the final answer, and a rejected edit receives error feedback and may be retried within the attempt budget. Failed edits and declared no-change proposals retain the parent score. For single-step revision we report mean and best-of-eight held-out scores, and for multistep revision the score of the development-selected harness (\pref{tab:single_step_families,tab:multistep_heldout_summary}). \pref{tab:validity} separates failed edits from zero-scoring harnesses, and \aref{app:scoring_sensitivity} examines sensitivity to the treatment of failed edits.

Unseen families exclude all SFT and RL families. \pref{tab:single_step_families} reports every evaluated family, its training exposure, and exclusions from the main figures. The main figures exclude five unseen families that both the 4B proposers and the teacher find hard to address with an algorithmic harness: knight\_swap, letter\_jumble, modulo\_grid, sokoban, and ab (marked $\dagger$). Over all 26 unseen families, mean scores are 0.282 for Base, 0.415 for SFT, 0.528 for Single-step RL, and 0.501 for the teacher.

The seed harness calls the solver once and extracts its answer. The fixed tool-loop baseline supplies Python execution for up to five turns and is shared across families. The teacher uses the same revision and scoring protocol and the same feedback reports.

Single-step figures include the seed baseline. Multistep curves plot retained-harness scores (\pref{eq:promote}), and the multistep columns of \pref{tab:variants} report the best proposal over rounds; the two scores differ by at most 0.002 in any family mean. \pref{fig:multistep_ood} compares both protocols on the shared unseen families, and \pref{tab:single_step_families,tab:multistep_ext24} give per-family results. Aggregate uncertainty is one standard error across families, computed after averaging repeated runs within each family, and per-family shading gives the range across runs.

\paragraph{Scope of revision comparisons.}
Single-step and multistep evaluation differ in proposal budget and prompting. Independent proposals use a branch instruction and regenerate the seed report. Sequential revisions share a fixed initial report and then use feedback from retained harnesses. Our comparisons of successive and independent revision do not isolate the contribution of feedback, because the protocols also differ in parent harness and candidate selection. \aref{app:multistep} details the Reasoning Gym family subsets, and \aref{app:qa_single_step} describes the corresponding QA comparison and proposal budgets.

Composition experiments compare standard prompts with prompts augmented by a helper-tool directive (\aref{app:composition_prompts}). Structural composition means that a harness integrates helpers into the solver's tool loop, and functional composition additionally requires a helper to contribute to the answer. \pref{tab:adjudication} reports the audit counts.

\subsection{SFT corpus and training}
\label{app:sft_details}
SFT uses first-round teacher revisions on the task families in \pref{tab:rg_training_families}, keeping those that improve seed performance and pass execution checks.

\begin{table}[htbp]
\centering
\small
\caption{\textbf{Task families used for SFT and RL on Reasoning Gym.} The unseen-family evaluation excludes every family used in either phase.}
\label{tab:rg_training_families}
\begin{tabular}{@{}lp{0.80\linewidth}@{}}
\toprule
Phase & Families \\
\midrule
SFT & \texttt{advanced\_geometry}, \texttt{binary\_alternation}, \texttt{bitwise\_arithmetic}, \texttt{boxnet}, \texttt{calendar\_arithmetic}, \texttt{color\_cube\_rotation}, \texttt{count\_bits}, \texttt{countdown}, \texttt{dice}, \texttt{fraction\_simplification}, \texttt{jugs}, \texttt{knights\_knaves}, \texttt{largest\_island}, \texttt{maze}, \texttt{mini\_sudoku}, \texttt{prime\_factorization}, \texttt{products}, \texttt{simple\_equations}, \texttt{syllogism}, \texttt{tower\_of\_hanoi}, \texttt{tsumego} \\
\addlinespace
RL & \texttt{codeio}, \texttt{maze}, \texttt{rotten\_oranges}, \texttt{simple\_geometry}, \texttt{tower\_of\_hanoi} \\
\bottomrule
\end{tabular}
\end{table}

Each collection context gives the teacher the seed's development score, outcome counts, and failed questions with trace excerpts, and the teacher independently revises a Python harness that exposes \texttt{solve(model, question)}. We execute each applied edit on the same set in a guarded sandbox. Accepted examples preserve completions verbatim, including reasoning, edits, and change manifests. \pref{tab:sft_settings} gives the collection budgets and corpus size. Each collection run samples eight first-round teacher proposals and retains one to six of them. Some collection runs add a directive asking the teacher to give the solver a Python execution tool; SFT training prompts omit this directive. Under the structural audit taxonomy (\aref{app:composition}), 207 of the 432 examples (48\%) are interpreter loops.

To enter the corpus, a candidate must satisfy $J(h';Q^{\mathrm{dev}})\geq\max\{J(h_0;Q^{\mathrm{dev}})+0.08,\,0.40\}$ and pass screens for answer echoing, duplicate code, and output-interface violations. Claude Opus 5 reviewers re-execute surviving candidates on development, held-out, and format-varied questions and inspect them for oracle access, memorization, and grader exploitation. No held-out question text enters the corpus.

Four-fold family-disjoint cross-validation over all 432 examples selects the duration of LoRA training. We then train on the full corpus for that duration and merge the adapters into the base model. \pref{tab:sft_settings} lists the optimization settings.

\paragraph{Limitations of supervised data.}
\label{app:sft_limitations}
Our Reasoning Gym SFT corpus contains only first-round revisions from a single teacher and is concentrated in interpreter loops and direct algorithmic solutions. Some collection runs also use explicit directives to encourage tool use. This supervision may limit the range of revision strategies learned during SFT, particularly for improving harnesses over successive rounds. \aref{app:limitations_future} discusses stronger supervision and alternative initializations.

\begin{table}[htbp]
\centering
\small
\caption{\textbf{Supervised data collection and training settings.}}
\label{tab:sft_settings}
\begin{tabular}{@{}p{0.40\linewidth}p{0.55\linewidth}@{}}
\toprule
Setting & Value \\
\midrule
\multicolumn{2}{l}{\textit{Data collection}} \\
Teacher & Qwen3.6-35B-A3B \\
Development instances per context & 25 (16 generated for \texttt{largest\_island} and \texttt{tsumego}) \\
Failed examples in the report & Up to 8, with question text and trace excerpts \\
Teacher proposals per context & 8 \\
Retained examples & 432 \\
\midrule
\multicolumn{2}{l}{\textit{Optimization}} \\
Adapter & LoRA, rank 32 and scaling parameter 32 \\
Target modules & All attention and MLP projections, including gated-delta-net projections \\
Context length & 32,768 tokens \\
Learning rate and schedule & $10^{-4}$, cosine decay, 10\% warmup \\
Batch size and accumulation & 1 example, 4 accumulation steps \\
Duration selection & Median best epoch from 4-fold cross-validation with disjoint families \\
Final training duration & Approximately 2.9 epochs \\
\bottomrule
\end{tabular}
\end{table}

\subsection{Single-step RL}
\label{app:rl_details}
Single-step RL uses a fixed pool of contexts pairing a task family, parent harness, and execution report. Feedback and reward instances share a family and use disjoint seeds. \pref{tab:rl_context_pool} lists contexts built from the seed and stronger parent harnesses.

\begin{table}[htbp]
\centering
\small
\caption{\textbf{Single-step RL context pool.} Training contexts by family and parent harness.}
\label{tab:rl_context_pool}
\begin{tabular}{@{}lrrr@{}}
\toprule
Family & Seed parent & Stronger parent & Total \\
\midrule
\texttt{tower\_of\_hanoi} & 100 & 100 & 200 \\
\texttt{codeio} & 100 & 100 & 200 \\
\texttt{rotten\_oranges} & 100 & 0 & 100 \\
\texttt{simple\_geometry} & 100 & 0 & 100 \\
\texttt{maze} & 40 & 0 & 40 \\
\midrule
Total & 440 & 200 & 640 \\
\bottomrule
\end{tabular}
\end{table}

The task-score term in \pref{eq:reward} is mean grader credit on the scoring set, including partial credit. The validity reward is
\begin{equation}
v_i=0.05\,\mathbf{1}[y_i\text{ parses}]
+0.10\,\mathbf{1}[h'_i\neq\bot]
+0.15\,\rho_{\mathrm{run}}(h'_i),
\label{eq:rg_validity_reward}
\end{equation}
where $\rho_{\mathrm{run}}$ is the fraction of scoring trials that complete. With unit task-score weight, a harness that completes every trial but answers incorrectly earns $0.30$. Failed edits earn only accrued validity credit. We execute and score the parent for no-change responses.

We sample revisions asynchronously with vLLM, apply each edit, execute the harness with the frozen solver, and compute task and validity rewards. Each sampled batch receives one policy update with group-normalized advantages and a loss averaged over response tokens. \pref{tab:rg_rl_settings} gives the sampling, optimization, and execution settings.

\begin{table}[htbp]
\centering
\small
\caption{\textbf{Single-step RL settings.} Solver execution settings are shared between training and evaluation.}
\label{tab:rg_rl_settings}
\begin{tabular}{@{}p{0.44\linewidth}p{0.51\linewidth}@{}}
\toprule
Setting & Value \\
\midrule
\multicolumn{2}{l}{\textit{Proposer sampling and optimization}} \\
Initialization & SFT checkpoint \\
Candidates per context ($G$) & 8 \\
Contexts per update & 40 \\
Learning rate & $10^{-6}$ \\
Updates per sampled batch & 1 \\
Training duration & 1 epoch over 640 contexts, 16 updates \\
KL penalty / entropy bonus & None / none \\
Sampling temperature & 0.6 \\
Prompt / response budgets & 12,288 / 16,384 tokens \\
Reported checkpoint & Update 10 \\
\midrule
\multicolumn{2}{l}{\textit{Harness execution}} \\
Solver & Frozen Qwen3.5-4B, non-thinking \\
Solver temperature & 0.7 \\
Solver response allowance & Minimum configured budget of 32,768 tokens \\
Scoring instances per context & 25, with partial credit \\
Timeout per scoring trial & 300 seconds \\
\bottomrule
\end{tabular}
\end{table}

Single-step RL denotes the checkpoint selected at update 10; rows marked \emph{later} use update 16. \aref{app:rl_ablations} reports additional training configurations with changes to the composition directive, training pool, and reward terms.

\subsection{Multistep RL}
\label{app:multistep_details}
Multistep RL on Reasoning Gym trains in two offline phases, each with the immediate-reward update of \sref{sec:rl}, and refreshes the revision states between them (\pref{tab:offline_revision_training}). Each revision receives the full response budget.

Each state's report comes from rescoring its harness on 25 canonical development questions and lists outcome counts and up to eight failing instances. The proposer receives the report, parent, task description, and revision instruction. All candidates share the context's scoring questions.

\begin{table}[htbp]
\centering
\small
\caption{\textbf{Offline training on successive revision states.}}
\label{tab:offline_revision_training}
\begin{tabular}{@{}p{0.20\linewidth}p{0.75\linewidth}@{}}
\toprule
Stage & Procedure \\
\midrule
Initialization & Start from the reported Single-step RL checkpoint (update 10). \\
First phase & Train for 9 updates on 62 stored parent harnesses with room for improvement. \\
State refresh & Advance each parent to its best sampled revision when it improves by at least 0.01, rescore all states on the same 25 development questions, and keep states at least 0.10 below their family's best score. \\
Second phase & Train for 15 updates on 246 states that pass the same filter, most drawn from earlier revision runs. The final checkpoint is reported as Multistep RL. \\
\bottomrule
\end{tabular}
\end{table}

The later-round-state experiment (\aref{app:later_round_states}) trains on a fixed mixture of contexts from SFT revision runs. Multistep RL and the later-round-state run share the inference protocol in \aref{app:protocol}. On HotpotQA, Multistep RL generates its training states online (\sref{sec:rl}).

\subsection{Training curves}
\label{app:train_curves}
\begin{figure}[htbp]
    \centering
    \includegraphics[width=\textwidth]{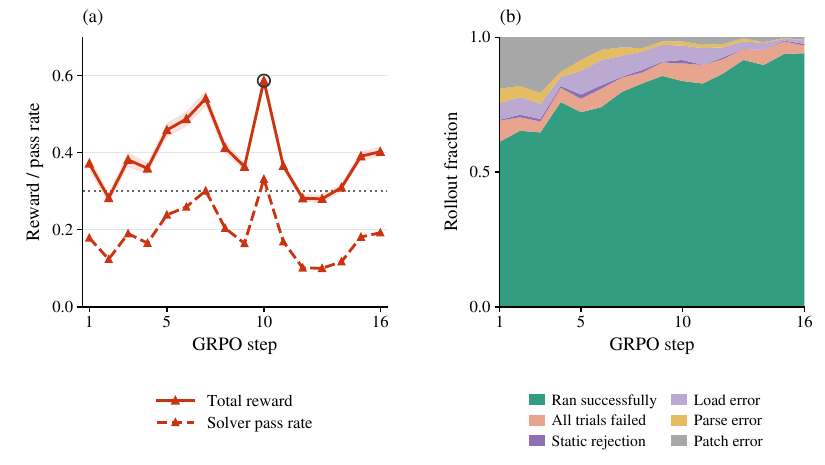}
    \caption{\textbf{Reasoning Gym RL training.} (a) Mean reward with one standard error, task score, and the validity-reward baseline. The open circle marks the reported checkpoint. (b) Rollout outcomes. The large variance comes from the prompt distribution, since a batch may mix tasks from different families. Training settings are in \aref{app:rl_details}.}
    \label{fig:rl_training_rg}
\end{figure}

\pref{fig:rl_training_rg} shows reward and rollout outcomes for single-step RL, with one unsmoothed point per GRPO update. \aref{app:rl_ablations} reports the additional training configurations and their evaluation results.

\ifdefined\arxivpreprint\else\clearpage\fi
\section{Additional Reasoning Gym Results and Analyses}
\label{app:rg_results}
\phantomsection
\label{app:table_conventions}
Methods and proposal accounting follow \aref{app:rg_training}. Single-step RL denotes checkpoint 10 unless specified otherwise. \pref{tab:appendix_methods} defines additional training configurations, and \aref{app:composition} defines composition rewards.

Each caption states whether its scores are held-out scores or the development scores used for selection. Missing scores appear as \texttt{--}, and blank cells denote evaluations that were not run.

\subsection{Single-step results}
\label{app:full_tables}

\begin{figure}[htbp]
    \centering
    \includegraphics[width=\textwidth]{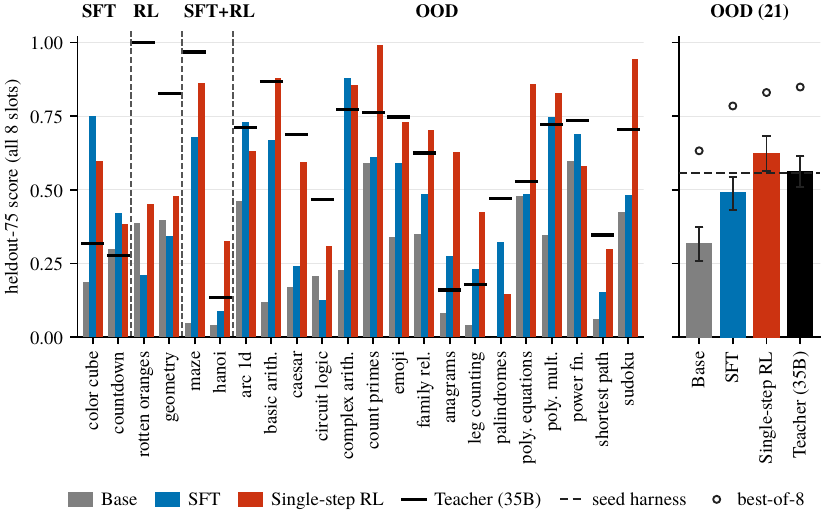}
    \caption{\textbf{Single-step revision by training exposure.} Bars give mean held-out scores over eight proposals, and black marks give the teacher. In the OOD summary panel, circles mark best-of-eight scores, the dashed line marks the seed, and error bars show one standard error across families. \pref{tab:single_step_families} lists every family and marks those excluded from the main comparison.}
    \label{fig:single_step_full}
\end{figure}

\pref{fig:single_step_full} groups single-step results by training exposure, and \pref{tab:single_step_families} reports every evaluated family, including those excluded from the main comparison. The mean score measures average proposal quality, the best-of-eight score measures performance with candidate selection, and the fixed tool-loop harness serves as a common baseline. On the 27 families with teacher results in the 28-family comparison, mean scores are 0.600 for Single-step RL and 0.567 for the teacher.

\begin{table}[htbp]
\centering
\small
\setlength{\tabcolsep}{3pt}
\caption{\textbf{Single-step revision by task family.} Entries report mean\,/\,best-of-eight scores across all eight attempted proposals on 75 held-out instances per family, and Tool loop gives the score of the fixed tool-loop harness. Single-step RL uses the selected checkpoint at update 10, \emph{later} is update 16, and Teacher is the 35B model. Families marked $\dagger$ are excluded from the 28-family comparison (\aref{app:protocol}). Teacher averages cover 32 families overall and 27 in the 28-family comparison.}
\label{tab:single_step_families}
\begin{tabular}{@{}llrrrrrr@{}}
\toprule
Family & Exposure & Tool loop & Base & SFT & \shortstack{Single-step\\RL} & \shortstack{Single-step\\RL (later)} & Teacher \\
\midrule
color\_cube\_rotation & SFT & 0.75 & 0.19\,/\,0.80 & 0.75\,/\,0.84 & 0.60\,/\,0.84 & 0.76\,/\,0.82 & 0.32\,/\,1.00 \\
countdown & SFT & 0.55 & 0.30\,/\,0.62 & 0.42\,/\,0.68 & 0.38\,/\,0.70 & 0.55\,/\,0.66 & 0.28\,/\,1.00 \\
codeio & RL & 0.32 & 0.24\,/\,0.29 & 0.17\,/\,0.29 & 0.27\,/\,0.34 & 0.30\,/\,0.36 & -- \\
rotten\_oranges & RL & 0.45 & 0.39\,/\,1.00 & 0.21\,/\,0.44 & 0.45\,/\,1.00 & 0.44\,/\,1.00 & 1.00\,/\,1.00 \\
simple\_geometry & RL & 0.53 & 0.40\,/\,0.68 & 0.34\,/\,0.56 & 0.48\,/\,0.58 & 0.55\,/\,0.57 & 0.83\,/\,1.00 \\
maze & SFT+RL & 0.95 & 0.05\,/\,0.36 & 0.68\,/\,0.93 & 0.86\,/\,0.95 & 0.87\,/\,0.97 & 0.97\,/\,0.99 \\
tower\_of\_hanoi & SFT+RL & 0.24 & 0.04\,/\,0.15 & 0.09\,/\,0.28 & 0.32\,/\,0.40 & 0.29\,/\,0.39 & 0.14\,/\,0.37 \\
ab$^\dagger$ & OOD & 0.39 & 0.08\,/\,0.47 & 0.28\,/\,0.56 & 0.16\,/\,0.43 & 0.33\,/\,0.52 & 0.21\,/\,1.00 \\
arc\_1d & OOD & 0.79 & 0.46\,/\,0.81 & 0.73\,/\,0.85 & 0.63\,/\,0.81 & 0.76\,/\,0.81 & 0.71\,/\,0.81 \\
basic\_arithmetic & OOD & 0.91 & 0.12\,/\,0.32 & 0.67\,/\,0.93 & 0.88\,/\,0.93 & 0.70\,/\,0.95 & 0.87\,/\,1.00 \\
caesar\_cipher & OOD & 0.97 & 0.17\,/\,0.55 & 0.24\,/\,0.91 & 0.59\,/\,0.99 & 0.68\,/\,0.98 & 0.69\,/\,0.96 \\
circuit\_logic & OOD & 0.13 & 0.21\,/\,0.57 & 0.12\,/\,0.25 & 0.31\,/\,0.77 & 0.28\,/\,0.77 & 0.47\,/\,0.68 \\
complex\_arithmetic & OOD & 1.00 & 0.23\,/\,0.79 & 0.88\,/\,1.00 & 0.86\,/\,1.00 & 0.96\,/\,1.00 & 0.77\,/\,0.97 \\
count\_primes & OOD & 0.99 & 0.59\,/\,0.69 & 0.61\,/\,0.99 & 0.99\,/\,1.00 & 0.80\,/\,1.00 & 0.76\,/\,1.00 \\
emoji\_mystery & OOD & 0.66 & 0.34\,/\,1.00 & 0.59\,/\,1.00 & 0.73\,/\,1.00 & 0.77\,/\,1.00 & 0.75\,/\,1.00 \\
family\_relationships & OOD & 0.75 & 0.35\,/\,0.63 & 0.48\,/\,0.75 & 0.70\,/\,0.80 & 0.59\,/\,0.81 & 0.62\,/\,0.83 \\
group\_anagrams & OOD & 0.91 & 0.08\,/\,0.63 & 0.27\,/\,0.85 & 0.63\,/\,0.95 & 0.69\,/\,0.92 & 0.16\,/\,1.00 \\
knight\_swap$^\dagger$ & OOD & 0.01 & 0.18\,/\,0.53 & 0.00\,/\,0.02 & 0.08\,/\,0.49 & 0.05\,/\,0.32 & 0.24\,/\,0.53 \\
leg\_counting & OOD & 0.44 & 0.04\,/\,0.12 & 0.23\,/\,0.56 & 0.43\,/\,0.57 & 0.55\,/\,0.60 & 0.18\,/\,0.37 \\
letter\_jumble$^\dagger$ & OOD & 0.29 & 0.38\,/\,0.57 & 0.23\,/\,0.31 & 0.34\,/\,0.41 & 0.34\,/\,0.44 & 0.48\,/\,0.65 \\
mahjong\_puzzle & OOD & 0.49 & 0.34\,/\,0.80 & 0.41\,/\,0.72 & 0.48\,/\,0.81 & 0.50\,/\,0.68 & 0.47\,/\,0.83 \\
modulo\_grid$^\dagger$ & OOD & 0.01 & 0.02\,/\,0.07 & 0.02\,/\,0.05 & 0.02\,/\,0.04 & 0.03\,/\,0.05 & 0.14\,/\,0.92 \\
n\_queens & OOD & 0.43 & 0.01\,/\,0.11 & 0.22\,/\,0.41 & 0.26\,/\,0.51 & 0.41\,/\,0.52 & 0.48\,/\,1.00 \\
palindrome\_partitioning & OOD & 0.43 & 0.00\,/\,0.00 & 0.32\,/\,1.00 & 0.15\,/\,0.49 & 0.27\,/\,0.76 & 0.47\,/\,1.00 \\
polynomial\_equations & OOD & 0.90 & 0.48\,/\,0.97 & 0.49\,/\,0.97 & 0.86\,/\,1.00 & 0.70\,/\,0.99 & 0.53\,/\,0.99 \\
polynomial\_multiplication & OOD & 0.85 & 0.35\,/\,1.00 & 0.75\,/\,0.96 & 0.83\,/\,0.92 & 0.75\,/\,0.91 & 0.72\,/\,1.00 \\
power\_function & OOD & 0.76 & 0.60\,/\,0.80 & 0.69\,/\,0.80 & 0.58\,/\,0.80 & 0.74\,/\,0.79 & 0.74\,/\,0.84 \\
propositional\_logic & OOD & 0.18 & 0.07\,/\,0.23 & 0.14\,/\,0.24 & 0.14\,/\,0.23 & 0.13\,/\,0.20 & 0.05\,/\,0.23 \\
puzzle24 & OOD & 0.84 & 0.93\,/\,0.95 & 0.95\,/\,0.99 & 0.94\,/\,0.99 & 0.90\,/\,0.99 & 0.94\,/\,0.97 \\
shortest\_path & OOD & 0.15 & 0.06\,/\,0.31 & 0.15\,/\,0.35 & 0.30\,/\,0.88 & 0.28\,/\,0.44 & 0.35\,/\,0.35 \\
sokoban$^\dagger$ & OOD & 0.00 & 0.03\,/\,0.20 & 0.01\,/\,0.04 & 0.03\,/\,0.12 & 0.03\,/\,0.08 & 0.17\,/\,0.68 \\
sudoku & OOD & 0.88 & 0.42\,/\,1.00 & 0.48\,/\,0.96 & 0.94\,/\,1.00 & 0.67\,/\,1.00 & 0.70\,/\,1.00 \\
zebra\_puzzles & OOD & 0.97 & 0.81\,/\,1.00 & 0.82\,/\,1.00 & 0.85\,/\,0.99 & 0.81\,/\,1.00 & 0.34\,/\,1.00 \\
\midrule
\textit{mean, all 33} & & 0.57 & 0.27\,/\,0.58 & 0.41\,/\,0.65 & 0.52\,/\,0.72 & 0.53\,/\,0.71 & 0.52\,/\,0.84 \\
\textit{mean, 28 families} & & 0.65 & 0.29\,/\,0.61 & 0.46\,/\,0.73 & 0.59\,/\,0.79 & 0.60\,/\,0.78 & 0.57\,/\,0.86 \\
\textit{mean, 21 OOD families} & & 0.69 & 0.32\,/\,0.63 & 0.49\,/\,0.78 & 0.62\,/\,0.83 & 0.62\,/\,0.81 & 0.56\,/\,0.85 \\
\textit{mean, 27 teacher families} & & 0.66 & 0.30\,/\,0.63 & 0.47\,/\,0.75 & 0.60\,/\,0.81 & 0.61\,/\,0.80 & 0.57\,/\,0.86 \\
\bottomrule
\end{tabular}
\end{table}

\pref{tab:validity} counts failed edits and zero-scoring harnesses over all attempted proposals, including unusable edits. An applied edit can receive no task credit, so a zero score alone does not establish that the code is invalid.

\begin{table}[htbp]
\centering
\caption{\textbf{Proposal validity by evaluation set.} Each method attempts eight proposals per family. A failed edit yields no scorable harness (\aref{app:protocol}), and a zero-scoring harness has a recorded score of zero on 25 development instances. The combined rate is the number of failed edits plus zero-scoring harnesses, divided by attempted proposals. The canonical, format-varied, and unseen (OOD) sets contain 15, 12, and 5 families, respectively.}
\label{tab:validity}
\footnotesize
\setlength{\tabcolsep}{4pt}
\renewcommand{\arraystretch}{1.12}
\begin{tabular}{@{}llrrrr@{}}
\toprule
Evaluation set & Method & Attempted & Failed edits & \shortstack{Zero-scoring\\harnesses} & \shortstack{Combined\\rate} \\
\midrule
Canonical & Base & 120 & 1 & 45 & 38\% \\
 & SFT & 120 & 0 & 27 & 23\% \\
 & Single-step RL & 120 & 0 & 18 & 15\% \\
 & Single-step RL (later) & 120 & 0 & 14 & 12\% \\
\midrule
Format-varied & Base & 96 & 4 & 29 & 34\% \\
 & SFT & 96 & 0 & 21 & 22\% \\
 & Single-step RL & 96 & 0 & 6 & 6\% \\
 & Single-step RL (later) & 96 & 0 & 8 & 8\% \\
\midrule
OOD & Base & 40 & 0 & 15 & 38\% \\
 & SFT & 40 & 1 & 13 & 35\% \\
 & Single-step RL & 40 & 0 & 3 & 8\% \\
 & Single-step RL (later) & 40 & 0 & 3 & 8\% \\
\bottomrule
\end{tabular}
\end{table}

\paragraph{Sensitivity to failed-edit scoring.}
\label{app:scoring_sensitivity}
\pref{tab:slot_rule} compares retaining the parent score, assigning zero, and excluding failed edits on eleven format-varied families and nineteen unseen families, which differ from the main comparison sets. Single-step RL exceeds the teacher on unseen families under all three rules. Excluding zero-scoring harnesses conditions the metric on success and can reverse this comparison, so our main results include zeros. On this nineteen-family evaluation, 30 of the teacher's 152 proposals score zero, compared with 19 for Single-step RL.

\begin{table}[htbp]
\centering
\caption{\textbf{Sensitivity to failed-edit scoring.} Columns give held-out mean scores when a failed edit keeps the seed score (Seed), scores zero (Zero), or is excluded (Applied). No edit counts proposals that produce no revised harness. Each family has eight proposals.}
\label{tab:slot_rule}
\footnotesize
\setlength{\tabcolsep}{4pt}
\renewcommand{\arraystretch}{1.12}
\begin{tabular}{@{}lrcccrccc@{}}
\toprule
 & \multicolumn{4}{c}{Format-varied (11 families)} & \multicolumn{4}{c}{Unseen (19 families)} \\
\cmidrule(lr){2-5}\cmidrule(lr){6-9}
Method & No edit & Seed & Zero & Applied & No edit & Seed & Zero & Applied \\
\midrule
Base & 0 & 0.254 & 0.254 & 0.254 & 5 & 0.304 & 0.274 & 0.304 \\
SFT & 0 & 0.415 & 0.415 & 0.415 & 0 & 0.400 & 0.400 & 0.400 \\
Single-step RL & 0 & 0.576 & 0.576 & 0.576 & 0 & 0.501 & 0.501 & 0.501 \\
Single-step RL (later) & 0 & 0.551 & 0.551 & 0.551 & 1 & 0.526 & 0.519 & 0.525 \\
Multistep RL & & & & & 1 & 0.534 & 0.528 & 0.535 \\
\shortstack[l]{Composition-prompt RL\\(update 10)} & 0 & 0.526 & 0.526 & 0.526 & 1 & 0.496 & 0.491 & 0.493 \\
\shortstack[l]{Harder-task RL\\(update 14)} & 0 & 0.600 & 0.600 & 0.600 & 0 & 0.561 & 0.561 & 0.561 \\
Teacher (35B) & 0 & 0.551 & 0.551 & 0.551 & 4 & 0.471 & 0.447 & 0.472 \\
\bottomrule
\end{tabular}
\par\smallskip
\begin{minipage}{\linewidth}\footnotesize\raggedright
All methods receive the same feedback reports. Format-varied results for Base, SFT, and Single-step RL come from the main evaluation run; other rows come from separate evaluation runs. Scoring rules are compared within each row. Blank cells denote evaluations that were not run.
\end{minipage}
\end{table}
\clearpage

\subsection{Multistep results}
\label{app:multistep}

Each question format covers 24 families under the revision and promotion protocol in \aref{app:protocol}. Twelve original families are supplemented with twelve families excluded from the main proposers' SFT and RL training. The additional families were selected partly by their earlier mean single-step scores (\pref{tab:multistep_family_selection}).

\begin{table}[htbp]
\centering
\small
\caption{\textbf{Selection of additional multistep evaluation families.} The criteria use the earlier single-step results.}
\label{tab:multistep_family_selection}
\begin{tabular}{@{}p{0.23\linewidth}p{0.72\linewidth}@{}}
\toprule
Selection group & Criteria \\
\midrule
First six families & Single-step RL outperforms SFT, which outperforms Base; the best of eight RL proposals reaches the fixed tool-loop baseline; the seed leaves room for improvement. \\
Remaining six & The performance-ordering requirement is relaxed to both trained proposers outperforming Base. \\
\bottomrule
\end{tabular}
\end{table}

The canonical and format-varied sets use different family lists (\pref{tab:multistep_ext24}). \pref{fig:multistep_ood} uses the 15 unseen families shared with the single-step comparison. We average repeated runs within each family before computing aggregate scores and standard errors. Each additional family has one run per method.

\begin{figure}[htbp]
    \centering
    \includegraphics[width=\textwidth]{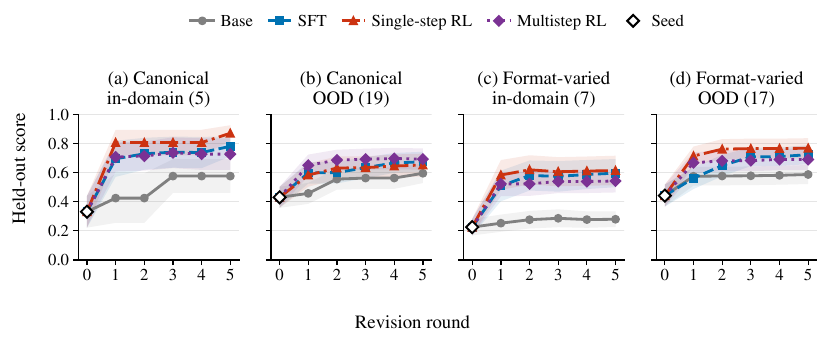}
    \caption{\textbf{Held-out performance over revision rounds.} Each point scores the harness that development-set selection retains after that round. Panels split families by question format and training exposure, with family counts in parentheses. A revision run keeps the seed until it promotes a proposal, and diamonds mark the seed score averaged over runs in each question format. Bands show one standard error across families, computed after averaging repeated runs within each family.}
    \label{fig:multistep_heldout}
\end{figure}

All three trained proposers finish above Base on the expanded held-out sets, with similar canonical scores and the highest format-varied score from Single-step RL (\pref{tab:multistep_heldout_summary}). \pref{fig:multistep_heldout} separates seen and unseen families across revision rounds.

\begin{table}[htbp]
\centering
\small
\setlength{\tabcolsep}{5pt}
\caption{\textbf{Held-out performance after five revisions.} Mean over 24 families per question format. Gains over Base are paired within families and reported as mean $\pm$ one standard error.}
\label{tab:multistep_heldout_summary}
\begin{tabular}{@{}lrrrr@{}}
\toprule
 & \multicolumn{2}{c}{Canonical} & \multicolumn{2}{c}{Format-varied} \\
\cmidrule(lr){2-3}\cmidrule(lr){4-5}
Method & Score & Gain over Base & Score & Gain over Base \\
\midrule
Base & 0.593 & & 0.498 & \\
SFT & 0.698 & $0.105\pm0.030$ & 0.688 & $0.189\pm0.058$ \\
Single-step RL & 0.699 & $0.106\pm0.044$ & 0.725 & $0.226\pm0.051$ \\
Multistep RL & 0.701 & $0.109\pm0.038$ & 0.649 & $0.150\pm0.051$ \\
\bottomrule
\end{tabular}
\end{table}

\pref{tab:multistep_ext24} reports development scores after rounds 1 and 5. On \texttt{caesar\_cipher}, the ordering of SFT and Single-step RL reverses between canonical and format-varied questions. \pref{fig:multistep_fam} shows the family-level revision curves across the original and additional families in both formats.

\begin{table}[htbp]
\centering
\footnotesize
\setlength{\tabcolsep}{2pt}
\renewcommand{\arraystretch}{1.08}
\caption{\textbf{Multistep revision by task family: canonical questions.} Columns r1 and r5 score the harness retained after rounds 1 and 5 on 25 development instances per family. All methods start from the same seed, so the Seed column is shared. Exposure labels follow \pref{tab:single_step_families}. Families below the dividing line form the additional evaluation set (\aref{app:multistep}) and have one run per method. Scores for the other families average repeated runs within each family. The format-varied results continue in the next panel.}
\label{tab:multistep_ext24}
\begin{tabular*}{\linewidth}{@{\extracolsep{\fill}}ll r rr rr rr rr@{}}
\toprule
Family & Exposure & Seed & \multicolumn{2}{c}{Base} & \multicolumn{2}{c}{SFT} & \multicolumn{2}{c}{Single-step RL} & \multicolumn{2}{c}{Multistep RL} \\
 & & & r1 & r5 & r1 & r5 & r1 & r5 & r1 & r5 \\
\midrule
basic\_arithmetic & OOD & 0.49 & 0.56 & 0.58 & 0.49 & 0.92 & 0.49 & 0.87 & 0.79 & 0.91 \\
bf & OOD & 0.00 & 0.00 & 0.00 & 0.04 & 0.04 & 0.00 & 0.00 & 0.00 & 0.04 \\
color\_cube\_rotation & SFT & 0.17 & 0.29 & 0.33 & 0.80 & 0.88 & 0.80 & 0.80 & 0.71 & 0.82 \\
count\_primes & OOD & 0.16 & 0.16 & 1.00 & 1.00 & 1.00 & 0.96 & 1.00 & 0.97 & 0.97 \\
countdown & SFT & 0.57 & 0.57 & 0.57 & 0.57 & 0.72 & 0.57 & 0.60 & 0.67 & 0.68 \\
cryptarithm & OOD & 0.47 & 0.47 & 0.52 & 0.47 & 0.47 & 0.47 & 0.49 & 0.48 & 0.48 \\
maze & SFT+RL & 0.08 & 0.08 & 0.64 & 0.96 & 1.00 & 0.64 & 0.88 & 0.84 & 0.97 \\
products & SFT & 0.48 & 1.00 & 1.00 & 0.92 & 1.00 & 1.00 & 1.00 & 1.00 & 1.00 \\
propositional\_logic & OOD & 0.16 & 0.16 & 0.16 & 0.16 & 0.16 & 0.16 & 0.16 & 0.16 & 0.17 \\
rotten\_oranges & RL & 0.20 & 0.20 & 0.44 & 0.20 & 0.56 & 1.00 & 1.00 & 0.48 & 0.55 \\
sudoku & OOD & 0.65 & 0.65 & 0.65 & 0.92 & 0.92 & 0.96 & 1.00 & 0.95 & 0.99 \\
word\_ladder & OOD & 0.06 & 0.06 & 0.06 & 0.06 & 0.06 & 0.10 & 0.10 & 0.07 & 0.07 \\
\midrule
ab & OOD & 0.48 & 0.64 & 0.64 & 0.48 & 0.56 & 0.48 & 0.48 & 0.48 & 0.48 \\
arc\_1d & OOD & 0.72 & 0.72 & 0.80 & 0.80 & 0.80 & 0.72 & 0.72 & 0.84 & 0.88 \\
caesar\_cipher & OOD & 0.00 & 0.00 & 0.64 & 0.16 & 0.88 & 0.00 & 0.07 & 0.88 & 1.00 \\
complex\_arithmetic & OOD & 0.69 & 0.81 & 1.00 & 1.00 & 1.00 & 1.00 & 1.00 & 1.00 & 1.00 \\
emoji\_mystery & OOD & 0.01 & 0.01 & 1.00 & 1.00 & 1.00 & 1.00 & 1.00 & 0.84 & 0.84 \\
family\_relationships & OOD & 0.52 & 0.52 & 0.56 & 0.72 & 0.72 & 0.68 & 0.80 & 0.56 & 0.72 \\
group\_anagrams & OOD & 0.56 & 0.56 & 0.56 & 0.56 & 0.56 & 0.56 & 0.96 & 0.96 & 0.96 \\
leg\_counting & OOD & 0.52 & 0.52 & 0.52 & 0.56 & 0.56 & 0.52 & 0.52 & 0.52 & 0.52 \\
mahjong\_puzzle & OOD & 0.76 & 0.76 & 0.80 & 0.76 & 0.76 & 0.76 & 0.76 & 0.76 & 0.76 \\
n\_queens & OOD & 0.80 & 0.80 & 0.80 & 0.80 & 0.80 & 0.80 & 0.80 & 0.80 & 0.80 \\
palindrome\_partitioning & OOD & 0.16 & 0.16 & 0.16 & 0.24 & 0.60 & 0.20 & 0.60 & 0.16 & 0.88 \\
polynomial\_multiplication & OOD & 0.80 & 0.80 & 0.80 & 0.84 & 0.96 & 0.80 & 1.00 & 0.88 & 1.00 \\
\textit{mean} & & 0.40 & 0.44 & 0.59 & 0.61 & 0.71 & 0.61 & 0.69 & 0.66 & 0.73 \\
\bottomrule
\end{tabular*}
\end{table}

\begin{table}[htbp]
\ContinuedFloat
\centering
\footnotesize
\setlength{\tabcolsep}{2pt}
\renewcommand{\arraystretch}{1.08}
\caption[]{\textbf{Multistep revision by task family: format-varied questions (continued).} Columns and evaluation protocol match the canonical panel. Families below the dividing line form the additional evaluation set.}
\label{tab:multistep_ext24_fmt}
\begin{tabular*}{\linewidth}{@{\extracolsep{\fill}}ll r rr rr rr rr@{}}
\toprule
Family & Exposure & Seed & \multicolumn{2}{c}{Base} & \multicolumn{2}{c}{SFT} & \multicolumn{2}{c}{Single-step RL} & \multicolumn{2}{c}{Multistep RL} \\
 & & & r1 & r5 & r1 & r5 & r1 & r5 & r1 & r5 \\
\midrule
basic\_arithmetic & OOD & 0.43 & 0.43 & 0.43 & 0.80 & 0.93 & 0.84 & 0.92 & 0.58 & 0.86 \\
codeio & RL & 0.28 & 0.28 & 0.28 & 0.33 & 0.35 & 0.29 & 0.40 & 0.31 & 0.38 \\
color\_cube\_rotation & SFT & 0.25 & 0.25 & 0.41 & 0.45 & 0.84 & 0.70 & 0.74 & 0.70 & 0.74 \\
count\_primes & OOD & 0.08 & 0.68 & 0.72 & 0.32 & 0.94 & 0.87 & 0.98 & 0.96 & 0.96 \\
countdown & SFT & 0.54 & 0.54 & 0.73 & 0.61 & 0.65 & 0.63 & 0.70 & 0.54 & 0.69 \\
maze & SFT+RL & 0.04 & 0.04 & 0.12 & 0.94 & 0.94 & 0.80 & 0.92 & 0.91 & 1.00 \\
propositional\_logic & OOD & 0.17 & 0.22 & 0.22 & 0.17 & 0.17 & 0.17 & 0.17 & 0.17 & 0.17 \\
rotten\_oranges & RL & 0.28 & 0.28 & 0.44 & 0.68 & 0.72 & 0.94 & 0.94 & 0.40 & 0.46 \\
simple\_geometry & RL & 0.04 & 0.25 & 0.32 & 0.36 & 0.58 & 0.56 & 0.58 & 0.59 & 0.64 \\
sudoku & OOD & 0.65 & 0.65 & 0.89 & 0.92 & 0.98 & 1.00 & 1.00 & 0.96 & 1.00 \\
tower\_of\_hanoi & SFT+RL & 0.10 & 0.10 & 0.10 & 0.11 & 0.30 & 0.22 & 0.38 & 0.35 & 0.46 \\
word\_ladder & OOD & 0.01 & 0.11 & 0.11 & 0.06 & 0.09 & 0.01 & 0.05 & 0.01 & 0.06 \\
\midrule
ab & OOD & 0.64 & 0.64 & 0.64 & 0.64 & 0.72 & 0.64 & 0.64 & 0.64 & 0.64 \\
arc\_1d & OOD & 0.72 & 0.72 & 0.84 & 0.72 & 0.76 & 0.72 & 0.72 & 0.72 & 0.84 \\
caesar\_cipher & OOD & 0.00 & 0.59 & 0.63 & 0.88 & 0.88 & 0.96 & 1.00 & 1.00 & 1.00 \\
complex\_arithmetic & OOD & 0.80 & 0.80 & 0.80 & 0.85 & 1.00 & 0.80 & 1.00 & 1.00 & 1.00 \\
emoji\_mystery & OOD & 0.01 & 1.00 & 1.00 & 0.01 & 0.84 & 1.00 & 1.00 & 0.84 & 0.84 \\
family\_relationships & OOD & 0.48 & 0.48 & 0.48 & 0.48 & 0.76 & 0.68 & 0.80 & 0.84 & 0.84 \\
group\_anagrams & OOD & 0.56 & 0.56 & 0.56 & 0.56 & 0.56 & 0.88 & 0.96 & 0.88 & 0.96 \\
leg\_counting & OOD & 0.52 & 0.52 & 0.52 & 0.52 & 0.56 & 0.52 & 0.72 & 0.56 & 0.56 \\
mahjong\_puzzle & OOD & 0.80 & 0.80 & 0.80 & 0.80 & 0.80 & 0.80 & 0.80 & 0.80 & 0.80 \\
n\_queens & OOD & 0.80 & 0.80 & 0.80 & 0.80 & 0.80 & 0.80 & 0.80 & 0.80 & 0.80 \\
palindrome\_partitioning & OOD & 0.20 & 0.20 & 0.20 & 0.20 & 1.00 & 0.64 & 0.64 & 0.28 & 0.28 \\
polynomial\_multiplication & OOD & 0.72 & 0.72 & 0.84 & 0.72 & 1.00 & 1.00 & 1.00 & 0.92 & 0.96 \\
\textit{mean} & & 0.38 & 0.49 & 0.54 & 0.54 & 0.72 & 0.69 & 0.74 & 0.66 & 0.71 \\
\bottomrule
\end{tabular*}
\end{table}

\begin{figure}[p]
    \centering
    \ifdefined\arxivpreprint
      \includegraphics[width=\textwidth]{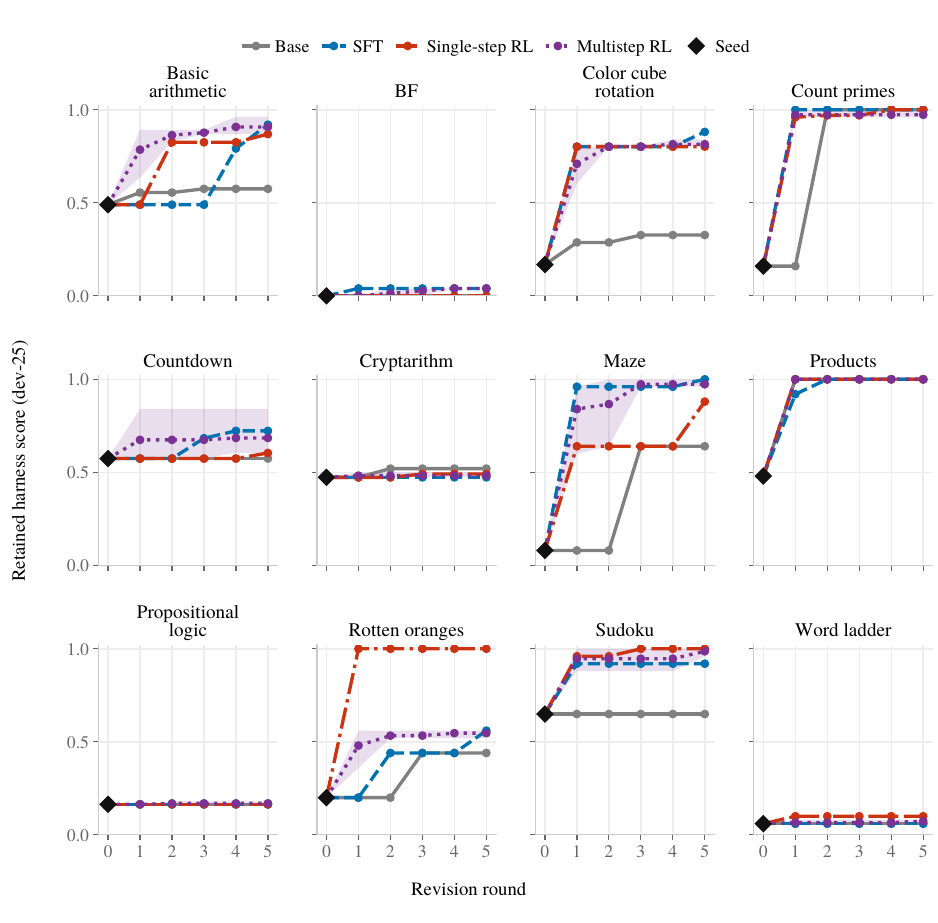}
    \else
      \includegraphics[width=\textwidth]{paper_figures/fig_multistep_refinement_fam.pdf}
    \fi
    \caption{\textbf{Development scores by family over revision rounds.} (a) Original families, canonical questions. Curves track the harness retained after each revision; diamonds mark the seed. Parts (b)--(d) continue on the following pages.}
    \label{fig:multistep_fam}
\end{figure}

\begin{figure}[p]
    \ContinuedFloat
    \centering
    \ifdefined\arxivpreprint
      \includegraphics[width=\textwidth]{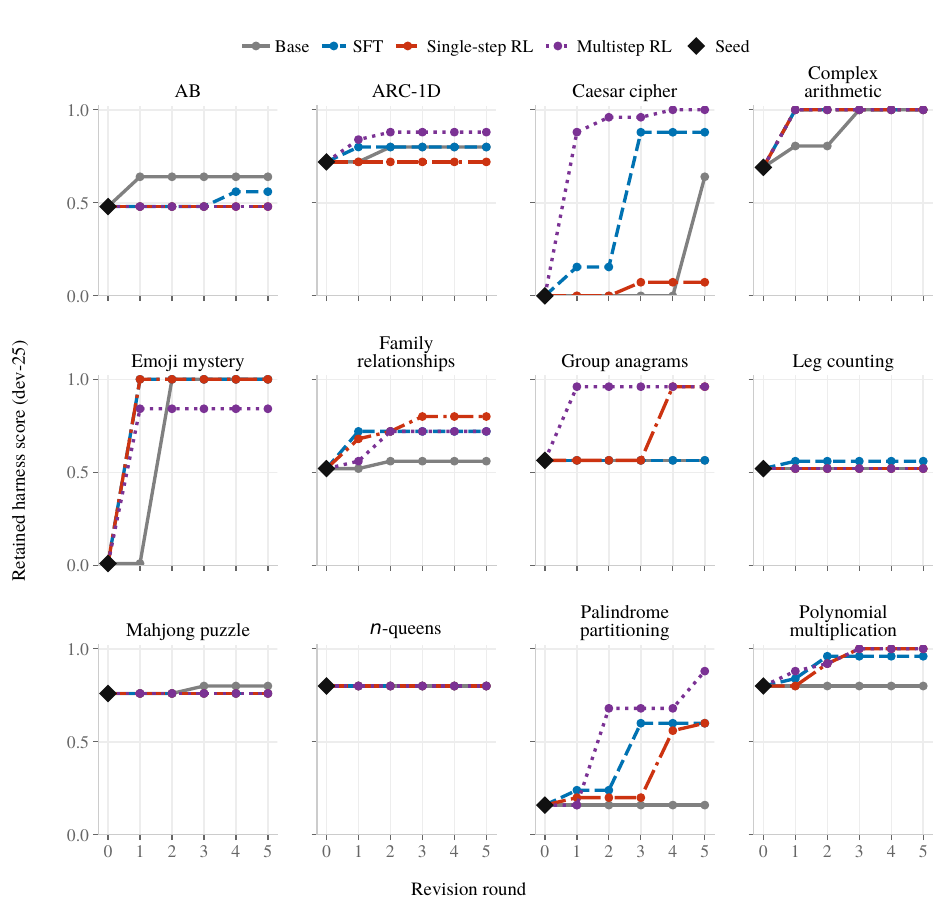}
    \else
      \includegraphics[width=\textwidth]{paper_figures/fig_multistep_refinement_fam_canonical_ext.pdf}
    \fi
    \caption[]{\textbf{Development scores by family (continued).} (b) Additional families, canonical questions. These families were selected as described in \aref{app:multistep} and have one run per method.}
    \label{fig:multistep_fam_ext}
\end{figure}

\begin{figure}[p]
    \ContinuedFloat
    \centering
    \ifdefined\arxivpreprint
      \includegraphics[width=\textwidth]{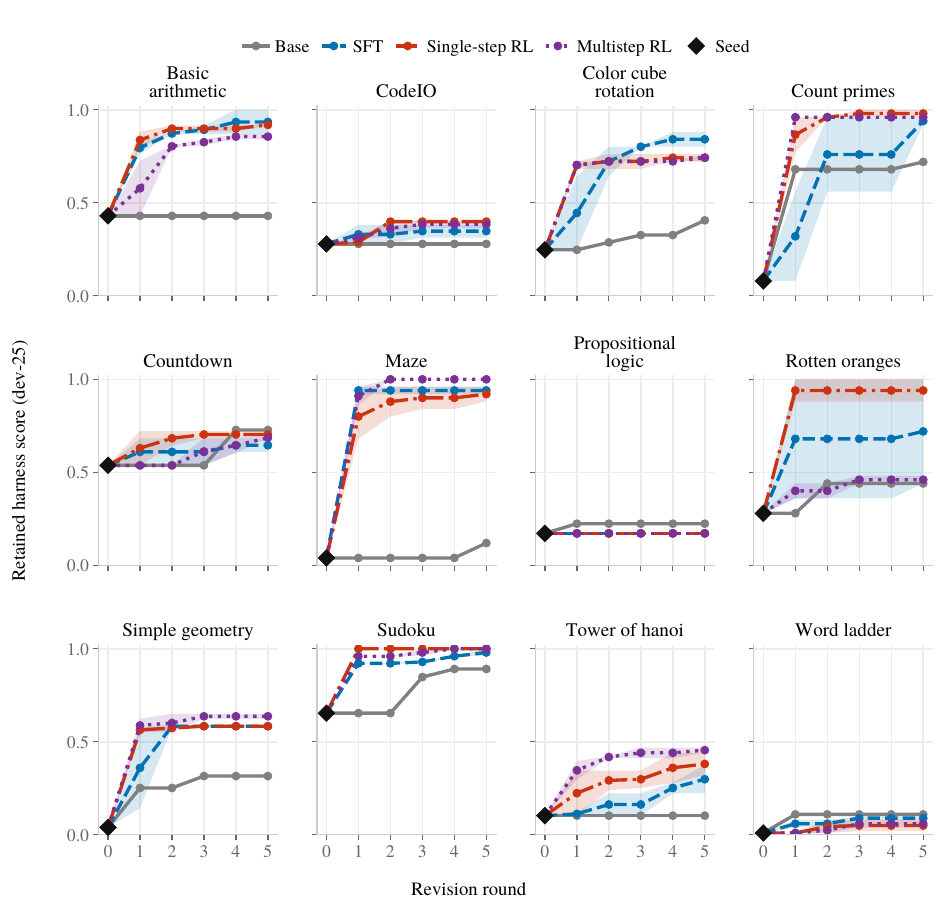}
    \else
      \includegraphics[width=\textwidth]{paper_figures/fig_multistep_refinement_fam_fmt.pdf}
    \fi
    \caption[]{\textbf{Development scores by family (continued).} (c) Original families, format-varied questions. Shading shows the minimum-to-maximum range across repeated runs within each family.}
    \label{fig:multistep_fam_fmt}
\end{figure}

\begin{figure}[p]
    \ContinuedFloat
    \centering
    \ifdefined\arxivpreprint
      \includegraphics[width=\textwidth]{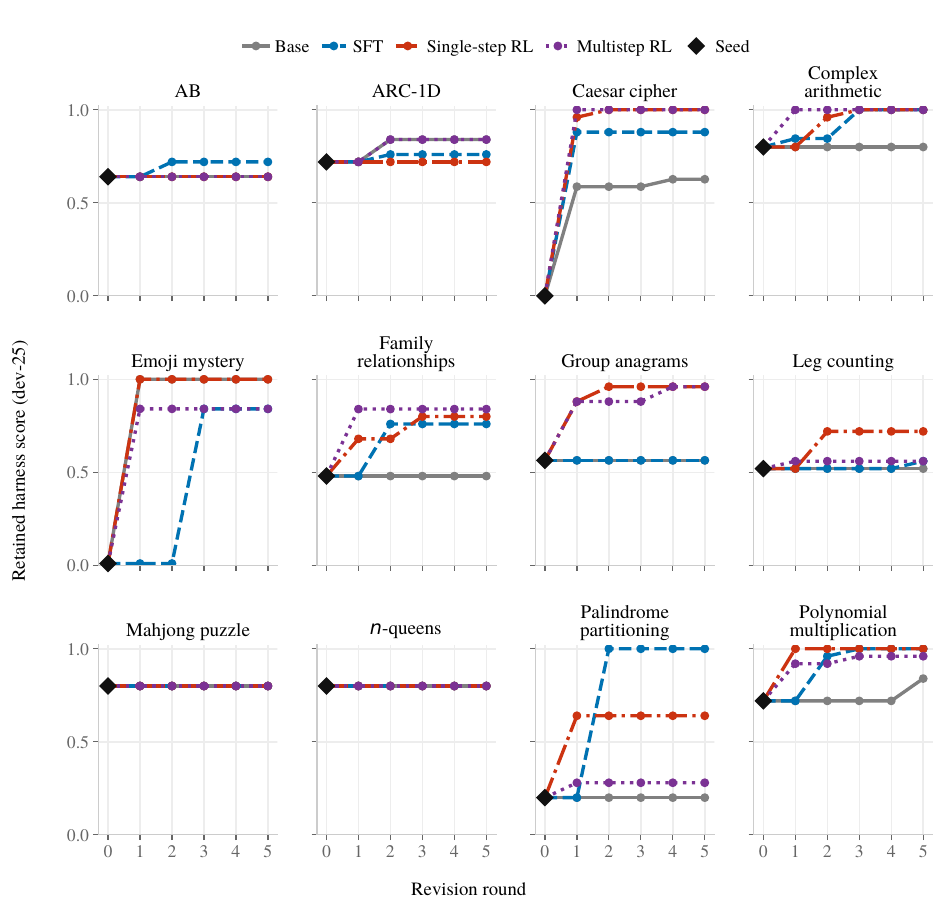}
    \else
      \includegraphics[width=\textwidth]{paper_figures/fig_multistep_refinement_fam_fmt_ext.pdf}
    \fi
    \caption[]{\textbf{Development scores by family (continued).} (d) Additional families, format-varied questions, with one run per method and family. Axes and method colors match parts (a)--(c).}
    \label{fig:multistep_fam_fmt_ext}
\end{figure}
\clearpage

\subsection{Training variants}
\label{app:rl_ablations}
Composition-prompt RL requests helper tools, while harder-task training changes the training pool and reward (\pref{tab:appendix_methods}). Both start from SFT and use $G=8$, a learning rate of $10^{-6}$, and no KL penalty, as in the main run. Because data, prompts, and reward terms can change together across configurations, each comparison measures the effect of the full configuration.

\begin{table}[htbp]
\centering
\footnotesize
\caption{\textbf{Additional training configurations.} Both variants use 20 contexts per update, compared with 40 in the main RL run.}
\label{tab:appendix_methods}
\begin{tabular}{@{}p{0.19\linewidth}p{0.37\linewidth}p{0.37\linewidth}@{}}
\toprule
Setting & Composition-prompt RL & Harder-task RL \\
\midrule
Training pool & 600 contexts combining low-scoring cases from the single-step RL pool (\aref{app:rl_details}) with additional data; includes 100 \texttt{maze} and 100 \texttt{sokoban} contexts & 694 contexts from seven families listed below \\
Prompt & Helper-composition directive & Canonical questions without a composition directive \\
Reward & \pref{eq:reward} & \pref{eq:p3_reward}, with revised validity terms and a structural-composition bonus \\
Training updates & 16 & 14 \\
Evaluated checkpoints & 10 and 16 & 14 \\
\bottomrule
\end{tabular}
\par\smallskip
\begin{minipage}{\linewidth}\footnotesize\raggedright
The harder-task pool contains \texttt{sokoban}, \texttt{modulo\_grid}, \texttt{ab}, \texttt{leg\_counting}, \texttt{rotten\_oranges}, \texttt{codeio}, and \texttt{simple\_geometry}. Prompt texts appear in \aref{app:composition_prompts}.
\end{minipage}
\end{table}

Composition-prompt RL produces fewer structurally composed harnesses at the later checkpoint, decreasing from 12 to 7 among 40 probe draws as the mean score increases slightly (\pref{tab:adjudication}). The score increase suggests that task reward can favor interpreter loops even when the prompt requests helper tools.

The harder-task pool contains families with room to improve over the fixed tool-loop harness and low expected scores for constant answers. Its reward includes a structural-composition bonus,
\begin{equation}
r_{\mathrm{hard}}=0.05\,\mathbf{1}[\text{loaded}]
+0.10\,\rho_{\mathrm{run}}+J(h';Q)
+0.15\,C(h'),
\label{eq:p3_reward}
\end{equation}
where $Q$ is the scoring set, $\rho_{\mathrm{run}}$ is the fraction of trials that complete, and $C(h')$ indicates whether the harness passes the structural composition screen. No recorded rollout passes the screen. The run therefore measures the combined effect of the harder pool and revised validity terms, with no reward from the composition bonus.

Harder-task RL scores above both evaluation runs of Single-step RL on the format-varied families (0.600 versus 0.576 and 0.506; \pref{tab:variants}), and composition-prompt RL exceeds the main run at its update-16 checkpoint (0.599). Harder-task RL finishes below Single-step RL after repeated revision on both development formats. Its nineteen-family evaluation includes training families and therefore measures a mixture of seen and unseen tasks.

\begin{table}[htbp]
\centering
\caption{\textbf{Training variants under single-step and multistep evaluation.} Single-step columns average all eight proposals on 75 held-out instances per family. Round 5 columns report the best proposal over five rounds on 25 development instances per family, using the original twelve-family sets.}
\label{tab:variants}
\footnotesize
\setlength{\tabcolsep}{4pt}
\renewcommand{\arraystretch}{1.12}
\begin{tabular}{@{}lrrrr@{}}
\toprule
 & \multicolumn{2}{c}{Single-step (75 held-out)} & \multicolumn{2}{c}{Round 5 (25 development)} \\
\cmidrule(lr){2-3}\cmidrule(lr){4-5}
Method & Format-varied & 19-family set & Canonical & Format-varied \\
\midrule
Base & & 0.304 & 0.496 & 0.398 \\
SFT & 0.463 & 0.400 & 0.645 & 0.625 \\
Single-step RL & 0.506 & 0.501 & 0.660 & 0.649 \\
Single-step RL (later) & & 0.526 & & \\
Multistep RL & & 0.534 & 0.637 & 0.618 \\
\shortstack[l]{Composition-prompt RL\\(update 10)} & 0.526 & 0.496 & & \\
\shortstack[l]{Composition-prompt RL\\(update 16)} & 0.599 & & & \\
\shortstack[l]{Harder-task RL\\(update 14)} & 0.600 / 0.595 & 0.561 & 0.640 & 0.643 \\
Teacher (35B) & 0.551 & 0.471 & & \\
\bottomrule
\end{tabular}
\par\smallskip
\begin{minipage}{\linewidth}\footnotesize\raggedright
Single-step format-varied results cover eleven families and use the same feedback reports as the main single-step comparison. SFT and Single-step RL values in this column come from a second evaluation run; the main run gives 0.415 and 0.576 (\pref{tab:slot_rule}). Replicate scores are separated by a slash. The nineteen-family set includes families used in harder-task training. Blank cells were not evaluated.
\end{minipage}
\end{table}

\paragraph{Later-round training states.}
\phantomsection
\label{app:later_round_states}
A separate run uses a fixed pool of 465 contexts from SFT revision sequences, comprising 240 first-round contexts, 155 later-round contexts, and 70 from revision runs that remain at the seed. Training starts from SFT, follows \aref{app:rl_details}, and runs for 22 updates; we evaluate checkpoint 16. The multistep run in \aref{app:multistep_details} refreshes states between phases.

The checkpoint reaches a round-five development score of 0.615, compared with 0.660 for Single-step RL, a paired difference of $-0.045\pm0.048$ across twelve families. Gains over rounds 2 through 5 are similar, providing no evidence that the fixed later-round states improve repeated revision.

\subsection{Harness composition}
\label{app:composition}

We audit generated harnesses for two properties: structural composition, where helpers are integrated into the solver's tool loop, and functional composition, where a helper contributes to the answer. \pref{tab:adjudication} reports both rates by evaluation group.

\begin{table}[htbp]
\centering
\caption{\textbf{Harness composition by prompt and training setting.}}
\label{tab:adjudication}
\footnotesize
\setlength{\tabcolsep}{3pt}
\renewcommand{\arraystretch}{1.12}
\begin{tabular*}{\linewidth}{@{\extracolsep{\fill}}lrrrl@{}}
\toprule
Method & $n$ & Structural & Functional & Most common class \\
\midrule
\multicolumn{5}{l}{\textit{Unseen families, without a directive}} \\
SFT & 39 & 0 & 0 & Interpreter loop (30) \\
Single-step RL & 40 & 0 & 0 & Interpreter loop (31) \\
\addlinespace[3pt]
\multicolumn{5}{l}{\textit{Unseen families, composition directive}} \\
SFT & 24 & 11 & 4 & Tool loop + helper (12) \\
Single-step RL & 32 & 12 & 3 & Interpreter loop (17) \\
\shortstack[l]{Composition-prompt RL\\(update 10, without directive)} & 32 & 0 & 0 & Interpreter loop (29) \\
\shortstack[l]{Composition-prompt RL\\(update 10)} & 32 & 7 & 5 & Interpreter loop (24) \\
\addlinespace[3pt]
\multicolumn{5}{l}{\textit{Composition-prompt RL, checkpoint probes with the directive}} \\
\shortstack[l]{Composition-prompt RL\\(update 10)} & 40 & 12 & 6 & Interpreter loop (27) \\
\shortstack[l]{Composition-prompt RL\\(update 16)} & 40 & 7 & 3 & Interpreter loop (33) \\
\addlinespace[3pt]
\multicolumn{5}{l}{\textit{Composition directive, comparison across training stages}} \\
Base & 38 & 0 & 0 & Algorithm (21) \\
SFT & 40 & 18 & 5 & Tool loop + helper (19) \\
Single-step RL & 40 & 15 & 5 & Interpreter loop (22) \\
\addlinespace[3pt]
\multicolumn{5}{l}{\textit{Guarded directive, helper with solver fallback}} \\
SFT & 40 & 0 & 0 & Algorithm (40) \\
Single-step RL & 40 & 0 & 0 & Algorithm (39) \\
\shortstack[l]{Composition-prompt RL\\(update 10)} & 40 & 0 & 0 & Algorithm (40) \\
\shortstack[l]{Composition-prompt RL\\(update 16)} & 40 & 0 & 0 & Algorithm (39) \\
\bottomrule
\end{tabular*}
\par\smallskip
\begin{minipage}{\linewidth}\footnotesize\raggedright
$n$ counts audited harnesses generated from development feedback. Structural and Functional count harnesses meeting the criteria in \aref{app:protocol}. Parentheses give counts for the most common class. Groups are separate evaluations, so counts are comparable within groups. Where specified, evaluations use the helper-composition directive. The guarded directive requests a helper with a solver fallback.
\end{minipage}
\end{table}

\ifdefined\arxivpreprint\else
\begin{figure}[htbp]
    \centering
    \includegraphics[width=\textwidth]{paper_figures/fig_class_heatmap.pdf}
    \caption{\textbf{Generated harness classes.} Each cell gives the percentage of audited harnesses in its row. The left block assigns each harness a primary class, and the right block gives structural and functional composition rates. A and B denote prompts without and with the composition directive.}
    \label{fig:class_heatmap}
\end{figure}
\fi

On the eleven-family format-varied set, the composition directive increases the composition-prompt model's score but lowers Single-step RL's score (\pref{tab:steering_scores}). In five-family probes, the composition-reward checkpoints also score lower with the directive.

\begin{table}[htbp]
\centering
\small
\caption{\textbf{Held-out scores with and without the composition directive.} Scores average eight proposals per family over eleven format-varied families, each evaluated on 75 held-out instances. All methods receive the same feedback reports, and all 88 proposals in each condition have recorded scores.}
\label{tab:steering_scores}
\begin{tabular}{@{}lrr@{}}
\toprule
Method & Without directive & With directive \\
\midrule
Single-step RL & 0.506 & 0.461 \\
\shortstack[l]{Composition-prompt RL\\(update 10)} & 0.526 & 0.559 \\
\bottomrule
\end{tabular}
\end{table}

A guarded directive requesting a deterministic helper with a solver fallback produces direct algorithmic solutions in 158 of 160 draws. No draw meets the structural or functional composition criteria (\pref{tab:adjudication}). Directive versions are specified in \aref{app:composition_prompts}.

\paragraph{Composition rewards.}
We compare tool-use and combined-reward variants, both initialized from SFT and trained on the composition-prompt pool with the task-specific helper directive. Each checkpoint is probed on five families with and without the directive, with eight proposals per family. A separate twelve-family evaluation measures repeated revision.

The added reward combines structural composition with a proxy for successful tool use. Let $F(h')$ average each scoring trial's task score multiplied by an indicator that a tool result reaches the solver, with $F(h')=0$ unless the structural screen passes. The reward is
\begin{equation}
r_{\mathrm{comp}}(h,y)=r(h,y)+w_{\mathrm{f}}F(h')+w_{\mathrm{s}}C(h'),
\label{eq:composition_reward}
\end{equation}
where $C(h')$ is the structural indicator in \pref{eq:p3_reward}. The tool-use reward uses $(w_{\mathrm{f}},w_{\mathrm{s}})=(0.3,0)$; the combined reward uses $(1.0,0.15)$. The audit assesses the helper's contribution to the answer separately from the reward proxy.

Under the tool-use reward, composed tool loops decline from 25 of 40 draws at initialization to one at checkpoint 14. At checkpoint 12, the combined reward retains 16 of 40 composed loops, but its round-five development score remains below Single-step RL (\pref{tab:composition_revision}). These probes do not isolate the effects of the reward terms.

\begin{table}[htbp]
\centering
\small
\caption{\textbf{Repeated revision after composition-reward training.} Round-five development scores on twelve canonical families, using the selection protocol in \aref{app:protocol} and 25 development instances per family.}
\label{tab:composition_revision}
\begin{tabular}{@{}llr@{}}
\toprule
Method & Checkpoint & Round-five score \\
\midrule
SFT & Initialization & 0.645 \\
Single-step RL & 10 & 0.660 \\
Tool-use reward & 14 & 0.624 \\
Combined reward & 12 & 0.624 \\
Combined reward & 20 & 0.630 \\
\bottomrule
\end{tabular}
\end{table}

Helper outputs reaching the solver in successful trials remain less common, and composed harnesses score below interpreter loops sampled during the same updates.

The checkpoint-12 inspection identifies parsing, tool communication, and answer extraction failures around correct helpers, including breadth-first search for \texttt{maze}, backtracking for \texttt{sudoku}, and a primality sieve for \texttt{count\_primes} (\pref{tab:composition_failures}).

\begin{table}[htbp]
\centering
\small
\caption{\textbf{Observed failures in composed harnesses.} Examples from the checkpoint-12 inspection.}
\label{tab:composition_failures}
\begin{tabular}{@{}p{0.22\linewidth}p{0.33\linewidth}p{0.38\linewidth}@{}}
\toprule
Component & Observed failure & Consequence \\
\midrule
Question parsing & The parser misses the goal marker. & Questions are routed to a fallback model call. \\
Answer extraction & The harness returns an unsolved grid. & The submitted answer is incomplete. \\
Tool communication & A tool message is malformed. & The tool interaction cannot execute. \\
\bottomrule
\end{tabular}
\end{table}

\FloatBarrier
\section{Multi-Hop QA Training and Evaluation}
\label{app:qa_details}

\subsection{Datasets and retrieval}

HotpotQA provides in-domain evaluation using 5.23 million Wikipedia abstracts from its full-wiki corpus. We sample disjoint sets of 2,000 training, 500 development, and 300 held-out test questions from its training split.

MuSiQue and 2WikiMultihopQA provide out-of-domain evaluation. We pool all supplied support and distractor paragraphs from their development splits (20 per question for MuSiQue-Ans and 10 for 2WikiMultihopQA) and deduplicate them by title and text. We partition each development split into disjoint development and held-out test sets, stratified by hop count for MuSiQue and question type for 2WikiMultihopQA. \pref{tab:qa_eval_data} lists the corpora and question counts. Development and test questions are disjoint from each other and from the 2,000-question training pool.

HotpotQA answers are compared by normalized exact match (lower-casing and removal of punctuation, articles, and extra whitespace). On MuSiQue and 2WikiMultihopQA, a prediction is also correct if it matches one of the gold answer's listed aliases, following the official 2WikiMultihopQA script.

\begin{table}[htbp]
\centering
\small
\caption{\textbf{Multi-hop QA evaluation data.} Retrieval corpora and question pools per benchmark.}
\label{tab:qa_eval_data}
\begin{tabular}{@{}lp{0.36\linewidth}rrl@{}}
\toprule
Benchmark & Retrieval corpus & Dev pool & Test & Split source \\
\midrule
HotpotQA & 5.23M Wikipedia abstracts (full-wiki) & 500 & 300 & train split, disjoint subsets \\
MuSiQue & 21,100 paragraphs pooled from the dev questions & 500 & 299 & dev split, hop-stratified \\
2WikiMultihopQA & 54,957 paragraphs pooled from the dev questions & 500 & 300 & dev split, type-stratified \\
\bottomrule
\end{tabular}
\end{table}

\subsection{RL training}
\label{app:qa_rl_details}

On HotpotQA, RL starts from the base proposer without an SFT stage and generates its training contexts online. For each context, we run the parent harness with traces on six feedback questions $Q^{\mathrm{fb}}$. The report lists the solver and retrieval calls, the final answer, whether it matches the gold answer, and which gold supporting documents were retrieved or missed. Each context also fixes 64 scoring questions $Q^{\mathrm{score}}$ on which all $G=8$ candidates are scored. Both sets come from the 2,000-question training pool, and each context's draw is seeded by its index. The proposer receives the parent harness, the report, and the revision instruction, and answers with a short diagnosis followed by one Python code block, without mutation hints.

The reward is exact match on the scoring questions plus a validity bonus,
\begin{equation}
r_i=\begin{cases}
J(h'_i;Q^{\mathrm{score}})+0.1 & h'_i\neq\bot\\
0 & \text{otherwise,}
\end{cases}
\label{eq:qa_reward}
\end{equation}
where $h'_i\neq\bot$ requires that the response contains a code block that parses, defines \texttt{run}, has at most 200 lines, and completes the scoring run. Questions on which the harness raises an exception or exceeds its call budgets count as incorrect inside $J$, so there is no separate partial-completion term.

\paragraph{Single-step RL.}
Every context uses the seed harness as parent with its own feedback and scoring draw. The pool holds 320 contexts, consumed once in 80 updates of four contexts each. We report the checkpoint after the final update as Single-step RL.

\paragraph{Multistep RL.}
Parents come from 32 revision runs that the evolving policy advances, so the training states come from successive revisions. Each run starts at the seed harness, and contexts are ordered by run, so each run is visited once every eight updates. When a run is revisited, the candidate from its previous context with the highest exact match on the scoring questions becomes the new parent, following the best-of-eight rule used at inference. A run keeps its parent if none of the eight candidates executes. Ten visits per run give 320 contexts and 80 updates per phase. The two policies share every other setting (\pref{tab:qa_rl_settings}), so the comparison isolates the source of the parent harness.

\begin{table}[htbp]
\centering
\small
\caption{\textbf{RL settings on multi-hop QA.} Single-step and Multistep RL share all settings unless noted. Training and evaluation use the same solver execution settings.}
\label{tab:qa_rl_settings}
\begin{tabular}{@{}p{0.44\linewidth}p{0.51\linewidth}@{}}
\toprule
Setting & Value \\
\midrule
\multicolumn{2}{l}{\textit{Contexts}} \\
Parent harness & Seed (Single-step); state of a revision run advanced by the policy (Multistep) \\
Feedback / scoring questions & 6 / 64, disjoint, from a 2,000-question training pool \\
Revision runs $\times$ depth (Multistep) & 32 $\times$ 10; commit by best-of-8 scoring exact match \\
Contexts per phase & 320 \\
\midrule
\multicolumn{2}{l}{\textit{Proposer sampling and optimization}} \\
Initialization & Base Qwen3-4B, thinking enabled (no SFT) \\
Candidates per context ($G$) & 8 \\
Contexts per update & 4 \\
Learning rate & $10^{-6}$ \\
Updates per sampled batch & 1 (clip ratio 0.2, token-mean loss) \\
Training duration & Single-step: 80 updates; Multistep: 80 updates \\
KL penalty / entropy bonus & KL-to-reference loss, coefficient $10^{-3}$ / none \\
Sampling temperature, top-$p$ & 1.0, 1.0 \\
Prompt / response budgets & 10,240 / 8,192 tokens \\
Reported checkpoint & Single-step: update 80; Multistep: update 80 \\
\midrule
\multicolumn{2}{l}{\textit{Harness execution}} \\
Solver & Frozen Qwen3-8B, non-thinking \\
Solver temperature, top-$p$, top-$k$ & 0.6, 0.95, 20 \\
Solver response allowance & 2,048 tokens (16,384-token context) \\
Retrieval & BM25 over 5.23M Wikipedia abstracts \\
Call budgets per question & 12 solver calls, 12 retrievals \\
Timeout & 120 s per solver call; 900 s per scoring run \\
\bottomrule
\end{tabular}
\end{table}

\subsection{Multistep evaluation}
\label{app:qa_eval_details}

All three benchmarks use the same two-hop retrieve-and-summarize seed. Each proposer performs four independent revision runs of $T=10$ rounds from this seed. In each round, each run draws six feedback questions from the development pool and executes its current harness on them with traces. The report lists traces, answer correctness, gold answers, and retrieved or missed supporting documents (\aref{app:prompt_feedback}). Sixty-four scoring questions from the same pool, disjoint from all feedback questions of the round, are shared by the four runs. Development questions can recur across rounds, while feedback and scoring questions remain disjoint within each round. Each run samples $G=8$ revisions and advances to the candidate with the highest scoring exact match, even when it scores below its parent. The Reasoning Gym protocol retains a candidate only when it improves the development score (\aref{app:revision_selection}).

A revision whose code does not parse, does not define \texttt{run}, exceeds 200 lines, or fails its scoring run scores zero, and a run keeps its harness only if all eight revisions fail. After each round, we score the four current harnesses on the full test set and report their mean. Dashed curves in the first three panels of \pref{fig:multihop_multistep} show the highest held-out score reached across all four runs up to each round. Random draws are seeded per round, so every proposer, including Base, sees the same feedback and scoring questions in the same round.

\paragraph{Sampling and execution.}
The proposer samples with temperature 1.0 and thinking enabled, with a response budget of 10,240 tokens. Base is the untrained Qwen3-4B under the same settings. Harnesses execute with the solver settings of \pref{tab:qa_rl_settings}. A scoring run is limited to 600 seconds and a test run to 1,800 seconds. The solver has a budget of 12 calls, and every retrieval returns at most 20 passages.

\begin{figure}[t]
  \centering
  \includegraphics[width=\textwidth]{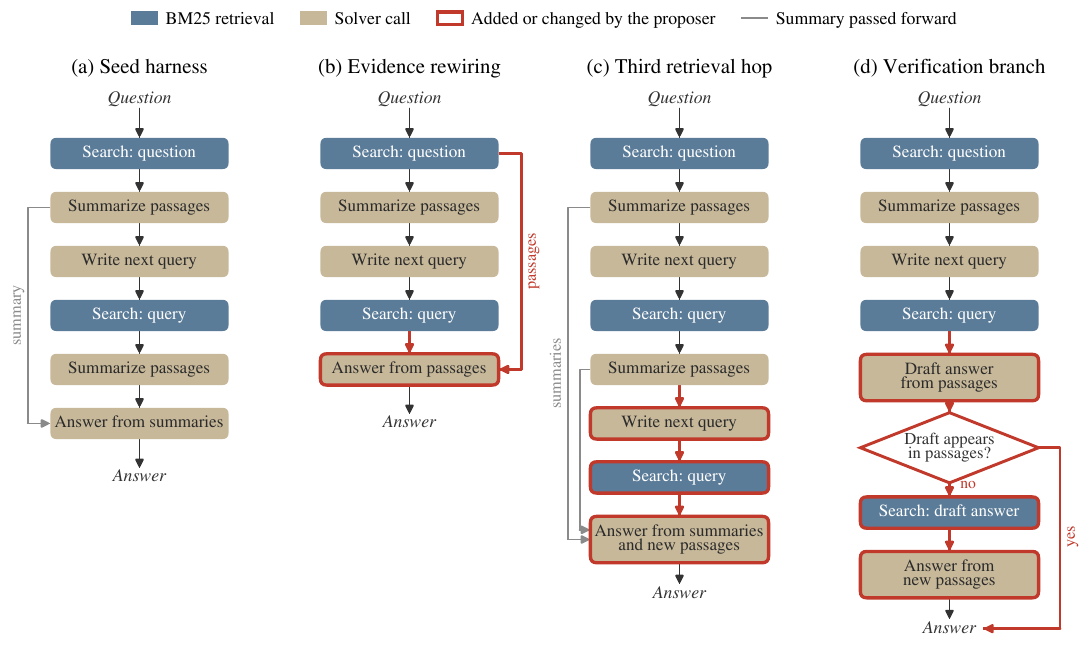}
  \caption{\textbf{Seed QA harness and three edits from adopted RL harnesses (abridged).} Red outlines mark additions and changes. (a) The seed retrieves twice and answers from solver summaries. (b) The answering call reads both hops' retrieved passages. (c) A third retrieval hop precedes answering. (d) The harness accepts a draft answer only if it appears in the retrieved passages. Otherwise the draft becomes a retrieval query, and the solver answers from the new passages.}
  \label{fig:qa_flowcharts}
\end{figure}

\subsection{Single-step evaluation}
\label{app:qa_single_step}

\paragraph{Proposal generation and scoring.}
For each dataset and RL checkpoint, we generate 80 independent proposals from the seed harness, using the round-one reports of the four revision runs and 20 samples per report. The pool size matches the 80 proposals that one revision run consumes in ten rounds. Each proposal receives the seed code and its execution feedback, produces one revised harness, and is scored by exact match on the held-out test set listed in \pref{tab:qa_eval_data}. Edits that fail to parse or compile receive zero. Identical programs share one execution, and all 80 proposals contribute to the reported statistics.

\pref{tab:qa_single_step} reports the unselected mean, the fraction strictly exceeding the seed, and the failed-edit rate. Oracle@8 is the expected maximum held-out score among eight proposals sampled uniformly without replacement from the 80. Best-of-80 is the maximum over the entire pool. Both oracle statistics select candidates by held-out score, so they describe the quality of the generated proposals and do not correspond to a deployable selection rule.

\begin{table}[t]
\centering
\small
\setlength{\tabcolsep}{3pt}
\caption{\textbf{Independent single-step QA revision.} Each row summarizes 80 proposals. The $>\!$seed column gives the fraction of proposals scoring strictly above the seed, and Failed gives the failed-edit rate. Oracle@8 and best-of-80 select on held-out scores.}
\label{tab:qa_single_step}
\begin{tabular*}{\linewidth}{@{\extracolsep{\fill}}llrrrrrr@{}}
\toprule
Dataset & Training & Seed & Mean & Oracle@8 & Best-of-80 & $>\!$seed & Failed \\
\midrule
HotpotQA & Single-step RL & 0.3900 & 0.4105 & 0.5399 & 0.5800 & 71.3\% & 1.3\% \\
 & Multistep RL & 0.3900 & 0.4131 & 0.5209 & 0.5833 & 77.5\% & 1.3\% \\
\addlinespace[3pt]
MuSiQue & Single-step RL & 0.1438 & 0.1467 & 0.2126 & 0.2408 & 58.8\% & 5.0\% \\
 & Multistep RL & 0.1438 & 0.1551 & 0.2034 & 0.2408 & 67.5\% & 2.5\% \\
\addlinespace[3pt]
2WikiMultihopQA & Single-step RL & 0.2500 & 0.3090 & 0.4746 & 0.5200 & 67.5\% & 1.3\% \\
 & Multistep RL & 0.2500 & 0.2748 & 0.4374 & 0.5433 & 58.8\% & 0.0\% \\
\bottomrule
\end{tabular*}
\end{table}

\paragraph{Comparison with successive revision.}
\pref{fig:multihop_multistep} compares these scores with the mean final score of four ten-round revision runs, whose selection uses only the scoring questions. Triangles show the highest held-out score reached across all ten rounds and four runs. Each run samples 80 candidates, matching the independent pool's proposal count, while the four-run mean and best-run score use 320 proposals in total. The best-run score is therefore not a matched-budget comparison with the best of 80 independent revisions. The two evaluations also distribute feedback differently. Independent proposals share four seed reports, and successive revision obtains feedback on intermediate harnesses. Seed performance is measured separately for the independent and successive evaluations; horizontal dashed lines in the bar panels show the independent-evaluation baseline.

\FloatBarrier
\clearpage
\section{Implementation Details}
\label{app:prompts}
\subsection{Structured harness revision}
\label{app:prompt_structured}

The following schematic excerpt describes the revision interface. It is not the complete production prompt. The proposer receives the task family and configuration, the current harness source and development score, and an execution report. It is asked to diagnose observed failures, make a targeted change without hard-coding answers, and preserve the harness entry point.

\begin{promptbox}{Revision input and output}
Task family and configuration: <TASK_DESCRIPTION>
Current harness and score: <PARENT_HARNESS_AND_DEV_SCORE>
Execution feedback: <REPORT>

Propose a targeted revision as SEARCH/REPLACE edits against the current source. 
Explain the change and identify the failures it is expected to fix and any likely regressions.
\end{promptbox}
Edits must match the current source and parse as a revised harness. The response includes a change manifest describing the edits and predicted effects; an unchanged harness must be declared as such. Invalid responses receive parsing or validation feedback and may be retried within the attempt budget in \pref{tab:rg_eval_settings}.

\label{app:prompt_response}

\subsection{Execution feedback}
\label{app:prompt_feedback}

For Reasoning Gym, the report provides development-set scores and instance outcomes, failed-example trace excerpts, changes since earlier revisions, and the retained harness's revision history. Independent proposals share the same parent and report. Their branch instruction directs different proposals toward distinct failure mechanisms, while successive revision obtains new feedback from the retained harness (\aref{app:protocol}). This distinction matters when comparing proposal budgets and adaptation over rounds.

Successive-state training supplies up to $K$ passing and $K$ failing examples, each with the question, harness answer, expected answer, and outcome (\aref{app:multistep_details}). Multi-hop QA supplies six development questions per revision run, with execution traces, answer correctness, gold answers, and retrieved or missed supporting documents (\aref{app:qa_details}).

\subsection{Composition directives}
\label{app:composition_prompts}

Composition experiments compare the standard revision request with one that additionally asks for a reusable computational helper integrated into the solver's tool loop. A more restrictive variant requests task-specific helpers instead of a generic code interpreter, along with checks of the resulting answer. These directives describe the desired harness structure; the structural and functional audits in \aref{app:composition} measure what the proposals actually achieve. The study's different training configurations and reward terms are specified in \pref{tab:appendix_methods} and \pref{eq:composition_reward}.

\clearpage
\section{Limitations and Future Work}
\label{app:limitations_future}

\paragraph{Teacher capability and proposer initialization.}
Our Reasoning Gym supervision is limited to the harness designs produced by a single 35B teacher. Stronger teachers could provide more diverse harness designs and multistep revision demonstrations, potentially better preparing the proposer for subsequent multistep RL. Another direction is to start RL from a model with prior training in harness design and revision. Our QA experiments already use direct RL without teacher demonstrations. Whether an initialization with more knowledge of harness design improves exploration and transfer remains an open question.

\paragraph{Agentic training of the proposer.}
Current training uses prescribed revision prompts and externally generated execution reports. The proposer does not choose its own code-inspection or execution steps before submitting a revision. Future work could train the proposer inside a tool-using harness, allowing it to inspect code, run targeted tests, compare candidate designs, and revise them using the resulting observations. This would extend training from producing edits to choosing how to investigate and improve a harness, and could help explore designs that are difficult to elicit through prompt directives alone.

\paragraph{Joint training and longer-horizon tasks.}
We keep the solver fixed during training, so our experiments do not examine whether learning to generate harnesses and learning to use them can reinforce each other. Concurrent work such as Ornith jointly trains scaffold generation and task solving \citep{ornith2026selfscaffolding,ornith2026selfimprovement}. An extension of harness learning could jointly optimize the proposer and solver while retaining the goal of generalizable adaptation. A key question is whether this training yields revision skills that transfer to unseen task families and remain useful as the solver changes, particularly on tasks requiring longer sequences of tool use and intermediate decisions.

\end{document}